\documentclass[10pt,journal]{IEEEtran}
\usepackage{amsmath,amsfonts}
\usepackage{algorithmic}
\usepackage{algorithm}
\usepackage{array}
\usepackage{tabularx} % tables that stretch to \columnwidth
\usepackage[caption=false,font=normalsize,labelfont=sf,textfont=sf]{subfig}
\usepackage{textcomp}
\usepackage{stfloats}
\usepackage{url}
\usepackage{verbatim}
\usepackage{graphicx}
\usepackage{capt-of}
\usepackage{cite}
\usepackage{anyfontsize}
\usepackage{amsmath,amssymb}
\usepackage{booktabs}
\usepackage{xcolor}
\usepackage[table]{xcolor}
\usepackage{enumitem}
\usepackage{hyperref}
\usepackage{multirow}

\IEEEaftertitletext{%
      \par
      \noindent
      \begin{minipage}{\textwidth}
          \centering
          \includegraphics[width=\textwidth]{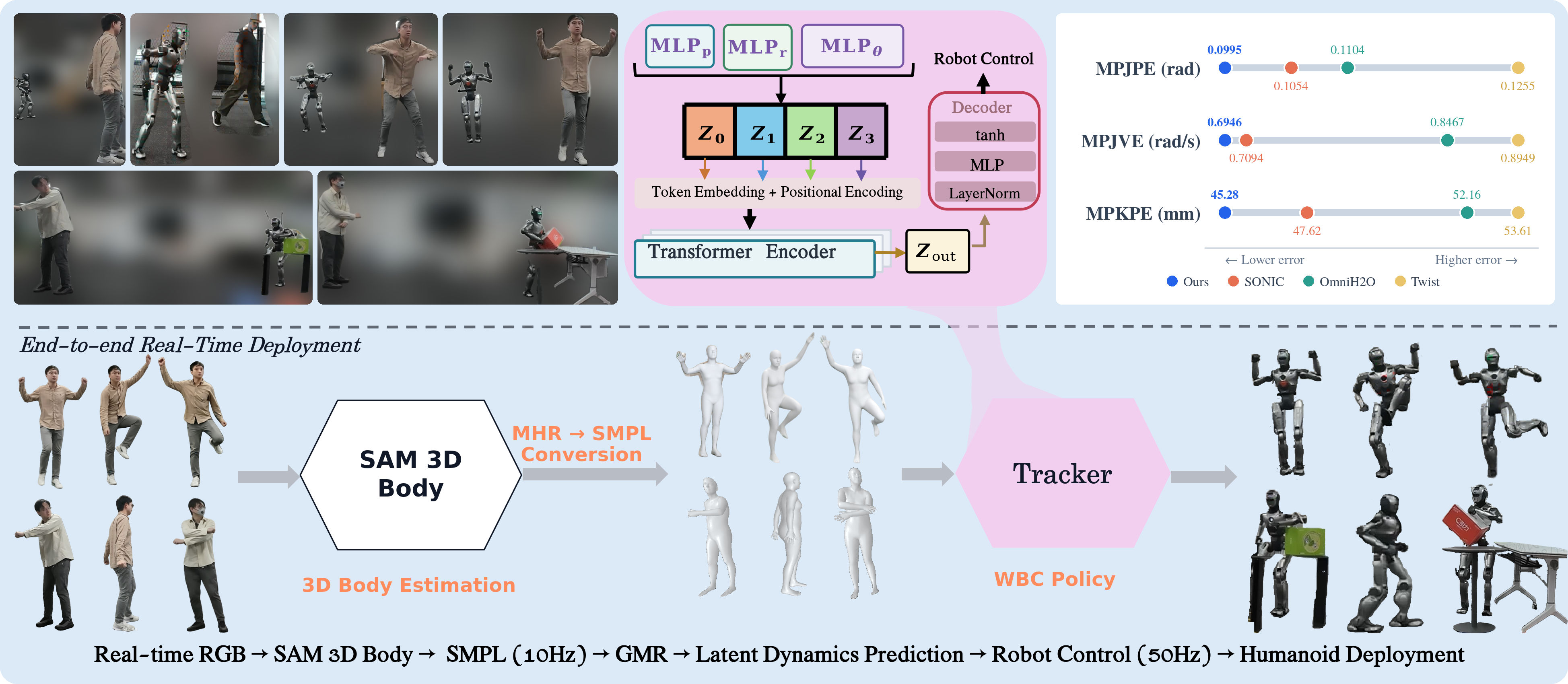}
          \captionof{figure}{Overview. Bottom: monocular RGB video is reconstructed, retargeted and extrapolated, producing 50-Hz control references. Top: imitation examples (left), the transformer tracker with frequency-domain conditioning (center), and tracking errors on an unseen dance sequence (right; lower is better).}
          \label{fig:overview}
      \end{minipage}
      \par
      \vspace{1.5em}
  }

\begin{document}

\title{MoViC: Real-Time Humanoid Whole-Body \textbf{C}ontrol from \textbf{Mo}nocular \textbf{Vi}deo Streams}

\author{%
Yimin~Han\textsuperscript{1,*},
Jiahui~Zhang\textsuperscript{1,*},
Wanyue~Li\textsuperscript{1,*},
Zeren~Luo\textsuperscript{1},
Yeke~Chen\textsuperscript{1},
Liu~Zhao\textsuperscript{1},
Peizhuo~Li\textsuperscript{2},
Xiaoyu~Ji\textsuperscript{1},
Shihao~Dong\textsuperscript{1},
Xuechao~Chen\textsuperscript{3},
Zhangguo~Yu\textsuperscript{3},
Annan~Tang\textsuperscript{4},
and~Peng~Lu\textsuperscript{1,\textdagger}%
\thanks{\textsuperscript{*}Equal Contribution. \protect\url{ymhan2023@connect.hku.hk}}%
\thanks{\textsuperscript{\textdagger}Corresponding author: \protect\url{lupeng@hku.hk}}%
\thanks{\textsuperscript{1}The University of Hong Kong, Hong Kong, China.}%
\thanks{\textsuperscript{2}National University of Singapore, Singapore.}%
\thanks{\textsuperscript{3}Beijing Institute of Technology, Beijing, China.}%
\thanks{\textsuperscript{4}EngineAI, Shenzhen, China.}}

\maketitle

\begin{abstract}
Real-time whole-body imitation from monocular video requires both a robust universal tracker and high-quality real-time reconstruction. We present MoViC, a framework that uses live RGB video as a control signal for humanoid robots without wearable tracking devices. The framework combines monocular human motion reconstruction, robot retargeting, and a transformer-based behavior foundation model.  The policy receives proprioceptive history, a single explicit next-step reference, and a compact frequency-domain motion representation that supplies temporal context without explicitly stacking future reference frames, which reduces the input dimension and network size. During deployment, our extrapolation method bridges the 10-Hz visual estimation with 50-Hz controller, enabling asynchronous visual processing and policy execution. Hardware experiments further demonstrate highly dynamic motion tracking and device-free teleoperation of bimanual and single-arm manipulation tasks from live monocular RGB input. Project page: https://movicproject.github.io/ .
\end{abstract}


\section{Introduction}
As humanoid robots increasingly enter everyday human environments, their behaviors are expected to become not only functional but also natural and human-like. Learning from large-scale human motion data provides a promising way to acquire such behavioral characteristics. However, achieving human-like whole-body control (WBC) on full-sized humanoid robots remains challenging. 
Traditional reinforcement learning (RL) approaches typically train task-specific policies with carefully designed rewards, but these policies transfer poorly across skills.
In contrast, extracting motion patterns from human motion data and learning a unified behavior foundation model (BFM) through motion tracking has emerged as an effective approach for humanoid WBC, enabling humanoid robots to acquire diverse whole-body behaviors directly from reference human motions.

Recent advances in large language models (LLMs) for natural language processing and vision-language-action (VLA) models suggest that data scale-up has become a key driver behind the emergence of generalist models with strong zero-shot generalization ability. Humanoid whole-body control can also benefit from the same principle. Imitation learning framework provide a promising way to leverage large-scale datasets for humanoid WBC~\cite{liao2026beyondmimic, tang2024humanmimic, peng2021amp, peng2018deepmimic}. However, some existing approaches, such as ASAP~\cite{he2025asap} and BeyondMimic~\cite{liao2026beyondmimic}, have limited generalization capacity across various motion modalities due to their model architectures or the limited scale and quality of their training data. These methods can reproduce highly dynamic in-domain motions, but fail to transfer zero-shot to unseen movements. In contrast, some methods, such as the TWIST series~\cite{ze2025twist, ze2025twist2} and UniTracker~\cite{yin2026unitracker}, demonstrate stronger generalization but tend to underfit highly dynamic behaviors. To solve this trade-off, recent studies have begun to adopt models that accommodate large-scale, diverse motion corpora and to scale up the datasets used for training, such as OmniH2O~\cite{he2024omnih2o}, and SONIC~\cite{luo2026sonic}, which achieve substantial progress in improving the learned policies' robustness in tracking diverse motions. Scaling the reference-motion dataset, however, raises a deeper question of how its statistics should be exploited. A large motion dataset—typically including motion capture (MoCap) data and video-reconstructed motions—defines a highly complex, multi-modal distribution over whole-body motions. 
Most prevailing methods imitate reference trajectories only implicitly through RL and treat the motion data as an unstructured collection of samples. Such an imitation framework clearly benefits from growing scale, but the rich structure of the underlying distribution, such as its behavioral modes and inter-motion relations, remains largely unexploited.
Beyond enlarging the motion scale, therefore, modeling the mathematical structure of the motion distribution offers a meaningful complement to building stronger behavior foundation models. Grouping similar motions provides the tracker with additional information for distinguishing different types of motion.

Moreover, existing popular humanoid real-time whole-body tracking systems can be roughly categorized into two types according to the teleoperation devices they use: MoCap-based~\cite{ze2025twist,qi2026humanoid,omnitrack2026} and VR-based~\cite{ze2025twist2, li2025clone}, as well as combinations of these two~\cite{luo2026sonic, he2024omnih2o}. Both of these solutions require expensive external hardware or operator-sensitive calibration, which restricts their deployment. MoCap-based methods typically depend on expensive, non-portable motion capture setups. VR-based methods require prior calibration that makes instant deployment impossible, and prolonged use of head-mounted displays (HMDs) can cause physiological burdens, such as visually induced motion sickness (VIMS). Vision-based solutions provide a practical alternative that preserves precise tracking ability while removing expensive and cumbersome wearable devices.
They require nothing beyond a low-cost RGB camera. As humanoid robots increasingly move into public life, daily interaction between humans and humanoids should not heavily rely on dedicated or specialized instruments. The visual interface should be an appropriate choice that lets whole-body control blend seamlessly into everyday human activities.

Video-driven WBC have been used, but mostly offline~\cite{mao2024learning,weng2025hdmilearninginteractivehumanoid}. The resulting policies do not follow a live operator. Video-based teleoperation instead requires online tracking of the operator's motion as it is estimated from the video stream. Curretnly, existing systems show that this is feasible with a single RGB camera~\cite{he2024learning,he2024omnih2o,fu2024humanplus,luo2026sonic}. However, they either track only sparse targets, which discards whole-body detail, or rely on additional sensing that complicates deployment. Besides, their monocular pose estimation is noisy and degrades under occlusion. High-quality motion reconstruction from video is typically too slow to keep pace with real-time control. SAM 3D Body~\cite{yang2026sam} targets robust full-body mesh recovery across diverse poses, but it runs at about 10~Hz, well below the control rate. To resolve this trade-off between reconstruction quality and speed, we propose an extrapolation method that bridges low-rate visual references and high-rate whole-body control, without requiring per-video training.

In summary, we propose MoViC framework. MoViC addresses the interface between visual motion estimation and whole-body control through the following three contributions:
\begin{enumerate}
    \item We develop MoViC, a monocular-video interface for real-time humanoid whole-body control. Human motion reconstruction and retargeting convert live RGB input into motion commands for a pretrained tracker, without wearable tracking devices or per-video policy training.
    \item We formulate a compact frequency-domain conditioning interface for a transformer-based behavior foundation model. A $\theta$ goal window supplies temporal motion context alongside proprioceptive history and a single explicit next-step reference, reducing the need for stacked future reference poses.
    \item We construct an online reference generator that combines extrapolation method with motion increments anchored to the latest measurement. It supplies 50-Hz references between approximately 10-Hz visual updates and compensates for reconstruction offsets shared by the current and predicted decoded states.
\end{enumerate}

\section{Related Work} 
\subsection{Whole-Body Humanoid Teleoperation}
Motion imitation enables physically simulated characters to acquire coordinated skills. PHC~\cite{luo2023perpetual} progressively expands a control policy for diverse motion tracking and fall recovery. Neural Categorical Priors~\cite{zhu2023neural} learn reusable vector-quantized skills. MaskedMimic~\cite{tessler2024maskedmimic} distills a transformer tracker into a conditional variational controller for partial motion constraints, while CALM~\cite{tessler2023calm} and PULSE~\cite{luo2024universal} learn motion-conditioned latent policies through adversarial imitation. But the physically simulated characters have different joint configurations and degree of freedom, compared with real humanoid robots. These methods have not yet
been demonstrated feasible on real humanoid robots.

For physical humanoids, ASAP~\cite{he2025asap} incorporates a residual action model learned from real-world data into simulation to reduce dynamics mismatch during policy refinement. However, the high cost of collecting real-world data makes such dynamics adaptation difficult to scale across diverse motion trajectories, potentially restricting the robot to a limited repertoire of learned motions. To enable a policy to track a broader range of reference motions, several works adopt similar teacher–student design. OmniH2O~\cite{he2024omnih2o}, HOVER~\cite{he2025hover}, and TWIST~\cite{ze2025twist} transfer privileged tracking skills to students operating with sparse targets, masked command modes, or restricted future references. UniTracker~\cite{yin2026unitracker} follows this paradigm while using a conditional variational student to model motion ambiguity under partial observations. However, these methods still fall short of achieving universal motion tracking ability.

To enable a single policy to track a broader range of reference motions, architectural improvements and temporal modeling of the input can enhance its representational capacity. Recent studies have made some progress toward universal motion tracking along these lines. HumanPlus~\cite{fu2024humanplus} and TWIST2~\cite{ze2025twist2} directly train history-conditioned tracking policies through reinforcement learning, using a decoder-only transformer and a convolutional encoder with an MLP, respectively.
GMT~\cite{chen2025gmt} and HoloMotion-1~\cite{chen2026holomotion} increase representational capacity through mixtures of experts, implemented as softly combined expert policies and sparsely routed transformer layers, respectively.
% GMT~\cite{chen2025gmt} increase representational capacity through mixtures of experts, implemented as softly combined expert policies. 
However, their performance is still bounded by how well the training data covers the space of motions. The out-of-distribution (OOD) motions are tracked poorly. Learned latent interfaces and generative models further broaden controller functionality. Exbody2~\cite{ji2024exbody2} extends whole-body tracking to expressive humanoid motion. SONIC~\cite{luo2026sonic} conditions control through shared latent representations and quantized codes, which unify heterogeneous motion commands. BeyondMimic~\cite{liao2026beyondmimic} and Heracles~\cite{tao2026heracles} combine tracking with motion synthesis, using guided latent diffusion for skill composition and state-conditioned trajectory generation for disturbance recovery, respectively. However, many of them rely on multi-stage training pipelines, such as teacher–student distillation, or latent-space pretraining. These make training costly, and hard to reproduce.

Moreover, fully exploiting the representational capacity of these high-capacity models with large parameters amount demands vast high-quality motion data. Training data combine public motion-capture datasets with video-reconstructed motions and self-captured data. Dataset scales reach hundreds or thousands of hours~\cite{luo2026sonic, chen2026holomotion}. As motion libraries grow, their structure can guide learning beyond providing reference targets and imitation rewards. However, few work mentioned above have considered the dataset structure. Some recent works, Humanoid-GPT~\cite{qi2026humanoid} and BumbleBee~\cite{wang2026experts} group motions for expert training and transformer distillation using Harmonic Motion Embedding and learned motion--text representations, respectively. But these methods both assign a single embedding to an entire sequence, so sequences containing multiple behaviors collapse into a mixed descriptor. Moreover, neither of them can be reused by the policy at deployment. A complementary use of motion structure is to encode local temporal context directly for policy conditioning.

Frequency-domain representations offer a structured way to encode this context. DeepPhase~\cite{starke2022deepphase} learns phase manifolds with a periodic autoencoder. FLD~\cite{li2024fld} extends this formulation with latent dynamics and multi-step consistency for periodic and quasi-periodic motion, enabling phase-based extrapolation. DFM~\cite{watanabe2025dfm} emphasizes local reconstruction and updated encodings to retain details in less periodic motions. This allows motions to be organized at the segment level rather than per sequence, and the same representation to be reused online for expert routing or continuous, adaptive sampling.

Video-driven deployment additionally requires handling reconstruction errors and the timing of reference updates. Several works learn humanoid behaviors from videos offline. Humanoid-X~\cite{mao2024learning} builds over 20 million humanoid poses from internet videos, HDMI~\cite{weng2025hdmilearninginteractivehumanoid} trains humanoid--object interaction policies from monocular videos, and GenMimic~\cite{ni2025generated} tracks motions reconstructed from generated videos. These methods process the videos before execution rather than following a live operator. Teleoperation instead requires online tracking, in which the robot imitates the operator's whole-body motion while it is being estimated from the video stream. Several systems achieve this visual interface with a single RGB camera. Although H2O~\cite{he2024learning} teleoperates a full-sized humanoid from a webcam, the robot's root velocity is still measured by a MoCap system during deployment. OmniH2O also supports an RGB camera, but tracks only the head and hand targets used for its VR interface~\cite{he2024omnih2o}. The sparse reference specifies only part of the operator's posture. VisualMimic~\cite{yin2025visualmimic} focuses on task-specific egocentric command generation, and requires a separate high-level policy, with its own reward design and privileged teacher, for each task. This task-dependent visual control is therefore unsuitable for vision-based teleoperation, which requires following arbitrary operator motions rather than executing a predefined task. HumanPlus~\cite{fu2024humanplus} shadows body and hand motions on a humanoid with WHAM~\cite{shin2024wham} and HaMeR~\cite{pavlakos2024reconstructing}. SONIC~\cite{luo2026sonic} accepts human video as a real-time command source. These systems show that video-based whole-body teleoperation is feasible, but they also expose some limitations. Firstly. the tracking ability is poor on the OOD highly dynamic motion references. Besides, the monocular pose estimation fails under large occlusions~\cite{he2024learning,fu2024humanplus}, and the WBC tracker has to tolerate the delayed reconstruction input~\cite{luo2026sonic}. The difficulties of pose estimations remain with the reconstruction models.

\subsection{Video Reconstruction}
Monocular human reconstruction offers a way to obtain motion references without body-worn tracking devices. VIBE~\cite{kocabas2020vibe} regresses pose and shape over time with an adversarial motion prior learned from motion-capture sequences. HMR~2.0~\cite{goel2023humans4d} estimates meshes with a transformer for human tracking in video. HybrIK~\cite{li2021hybrik} solves joint swings from 3D keypoints and regresses only 1-DoF twists, yielding pixel-aligned and kinematically valid meshes. But it is sensitive to keypoint errors. WHAM~\cite{shin2024wham} and GVHMR~\cite{shen2026gvhmr} further recover world-grounded motion, using camera angular velocity with contact-aware refinement and gravity-view coordinates, respectively. However, such models are largely trained on laboratory captures or pseudo-labeled data and often fail on challenging poses and under severe occlusion. Moreover, they rely on temporal context rather than per-frame inference. GVHMR in particular reconstructs entire sequences offline, which incurs latency and suffers leads to severe degradation under strong camera jitter. SAM 3D Body~\cite{yang2026sam} targets these failure cases. It is trained on annotations from a data engine that selects rare poses and imaging conditions, and represents the full body with the Momentum Human Rig, which decouples skeletal pose and body shape. It outperforms prior single-image methods and is competitive with video-based approaches. Because it requires only a single image, it can process a live stream frame by frame, which suits real-time teleoperation.

Our proposed MoViC combines temporal reference conditioning and online reference generation for whole-body control from live RGB video. MoViC implements SAM 3D Body for per-frame human reconstruction, and extrapolation for generating 50-Hz references from available visual estimates to align the reference trajectory sampling rate with the control rate. We combine existing reconstruction capabilities with structured motion conditioning, where compact motion representations are fed to a task-agnostic WBC tracker.

\section{System Overview}
As illustrated in Fig.~\ref{fig:overview}, MoViC drives humanoid whole-body control from live monocular RGB video through an online video-to-motion module, SEG, which reconstructs human motion with SAM 3D Body~\cite{yang2026sam} at approximately 10~Hz, and then retargets and extrapolates it to the robot into reference motions at 50~Hz. A transformer-based behavior foundation model then tracks these references by combining proprioceptive history with an explicit next-step reference and a frequency-domain $\theta$ goal.

\subsection{Data Curation}
We use a combination of AMASS \cite{mahmood2019amass}, LAFAN1 \cite{harvey2020robust}, Motion-X++ \cite{lin2023motionx}, OMOMO \cite{li2023object}, and self-captured motion data to train a BFM as our robust tracking policy. Because these datasets contain human motion sequences, we retarget them to the robot to form the reference-motion library for training a motion tracker. 

\noindent\textit{Motion Retargeting:} %\label{sec:retargeting}
When constructing the reference-motion library, the motions are converted to a common human-body representation~\cite{SMPL-X:2019}, temporally resampled, and retargeted to the EngineAI PM01 through parallel differentiable kinematic optimization. Let $\mathbf{y}_{i,t}$ denote the position of human body keypoint $i$, and let $\mathbf{p}_{i,t}$ denote the position of its corresponding robot link. We solve

\begin{equation}
\begin{aligned}
\min_{\mathbf{q},\mathbf{T},s,\{s_{ij}\}}\quad
&\lambda_{\mathrm g}\sum_{t,i}
  \|\mathbf{p}_{i,t}-s\mathbf{y}_{i,t}\|_2^2 \\
&+\lambda_{\mathrm l}\sum_{t,(i,j)\in\mathcal E}
  \|s_{ij}\Delta\mathbf{p}_{ij,t}
       -s\Delta\mathbf{y}_{ij,t}\|_2^2
\\
&+\lambda_{\mathrm r}\sum_{t,k\in\mathcal K}
  d_{\mathrm{SO}(3)}^2
  \!\left(\mathbf{R}_{k,t},\mathbf{R}^{\mathrm h}_{k,t}\right)
+\mathcal R,
\end{aligned}
\label{eq:gpu_retarget}
\end{equation}

\noindent where $\Delta\mathbf{p}_{ij,t}=\mathbf{p}_{i,t}-\mathbf{p}_{j,t}$, with $\Delta\mathbf{y}_{ij,t}$ defined analogously; $\mathcal E$ denotes the set of kinematically connected keypoint pairs. The set $\mathcal K$ contains links with orientation targets, $\mathbf{R}_{k,t}$ and $\mathbf{R}^{\mathrm h}_{k,t}$ are the corresponding robot and human orientations, and $d_{\mathrm{SO}(3)}$ denotes their distance on the rotation manifold. The global and pairwise scales $s$ and $s_{ij}$ account for morphology differences, while $\mathcal R$ contains regularization terms for joint-position and joint-velocity limits and temporal smoothness. The global term anchors the robot keypoints to the reference trajectory, the local term preserves the relative configuration of connected body parts, and the rotation term aligns the selected link orientations. We jointly optimize the root poses, joint configurations, and scale factors for batches of variable-length clips using GPU-parallel forward kinematics and automatic differentiation; padded frames are excluded from the objective by a validity mask.

\subsection{Universal Motion Tracker}
\subsubsection{Overview}
In this section, a scalable training framework for a universal motion tracker is proposed. Unlike other trackers that use the explicitly stacking future reference frames, we innovatively integrate the future information into frequency-domain motion representations, and design the actor architecture accordingly. The proposed framework, MoViC, integrates two coupled components: (i)~a pretrained motion encoder, built on latent dynamics, compresses the reference motions into a compact periodic parameterization $\theta$ and (ii)~a tracking policy incorporates $\theta$ into a Transformer actor. The actor receives ten frames of proprioceptive history, an explicit next-step reference, and a $\theta$ goal window. During training, this $\theta$ window conveys information about ten future reference frames. The explicit target specifies the immediate tracking objective, while the frequency-domain representation supplies additional motion context.

Most motions are relatively simple and can be learned with the next reference frame only, while more challenging motions may benefit from a longer future context. We therefore augment the explicit next-frame reference with a compact $\theta$ token that summarizes the next ten reference frames. Compared with explicitly providing ten or more future reference frames, this formulation reduces the input dimensionality, and thus supports learning with a smaller network, enabling the policy to acquire a broad repertoire of motions spanning different levels of difficulty.

\begin{figure*}[t]
\includegraphics[width=0.95\textwidth]{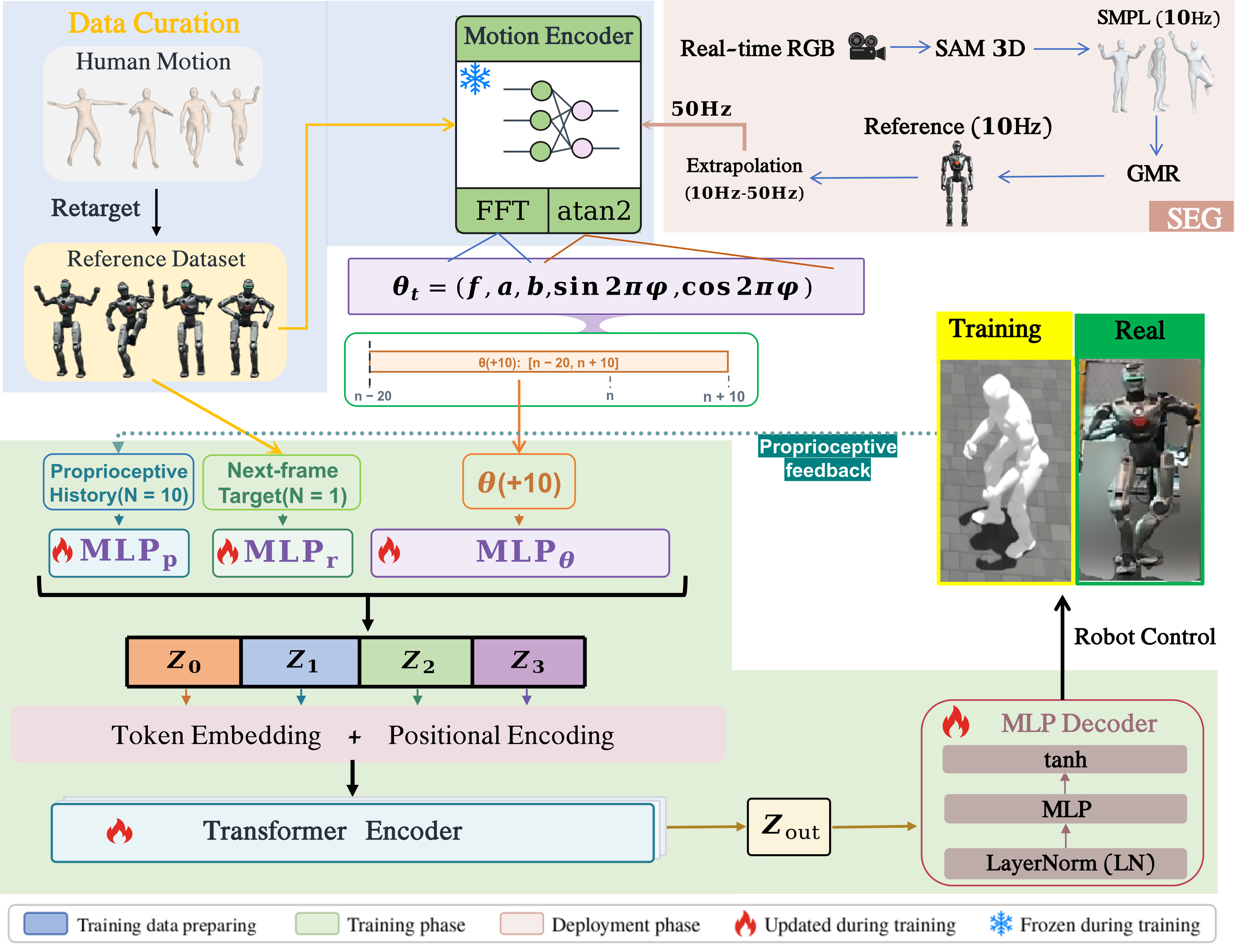}
    \centering
   \caption{Overview of the proposed whole-body motion-tracking framework. (a) A frozen DFM encoder maps reference motion windows to phase-augmented frequency-domain representations. (b) The transformer actor integrates proprioceptive history, a single next-control-step reference target, and $\theta$ token through self-attention. The output at the learnable query position is mapped to 24 joint position targets for PD control. (c) At deployment, The reference motions are provided together by the \textbf{SEG} module that is composed of SAM 3d Body reconstruction, latent-dynamic extrapolation and motion retargeting (GMR)~\cite{joao2025gmr}. The extrapolation generates intermediate references for 50-Hz policy execution from an approximately 10-Hz motion stream. (d) The policy is optimized using PPO with a privileged critic that receives privileged state history and the next-step reference target, but no $\theta$ codes. The dashed branch is used only during training.}
   \label{pics:net}
   %\vspace{-1.0em}
\end{figure*}

\subsubsection{Problem Formulation}
Whole-body motion tracking for the 24-DoF EngineAI PM01 is formulated as a partially observable Markov decision process (POMDP), $\mathcal{M}=(\mathcal{S},\mathcal{O},\mathcal{A},\mathcal{P},\Omega,\mathcal{R},\gamma)$. Here, $\mathcal{S}$, $\mathcal{O}$, and $\mathcal{A}$ denote the state, observation, and action spaces; $\mathcal{P}$ and $\Omega$ are the transition and observation models; and $\mathcal{R}$ and $\gamma$ denote the reward function and discount factor. The underlying state includes the robot, environment, and reference-motion process, whereas the policy receives only proprioceptive measurements and reference-derived features.

We optimize the policy using proximal policy optimization (PPO)~\cite{schulman2017proximal} to maximize the expected discounted return,
\begin{equation}
J(\pi)=\mathbb{E}_{\pi}\!\left[\sum_{t=0}^{T-1}\gamma^t R_t\right],
\end{equation}
where $T$ is the episode length, $\gamma=0.99$, and $R_t$ is the tracking reward defined in Section~\ref{sec:reward}.

\subsubsection{Asynchronous Actor-Critic}
Training uses an asymmetric actor-critic architecture. The transformer actor maps $\mathbf{o}^{\mathrm{act}}_t$ to an action distribution, whereas the critic estimates the discounted return from ten privileged state observations and ten future references:

\begin{equation}
V_{\psi}\!\left(
\mathbf{o}^{\mathrm{priv}}_{t-N+1:t},
\mathbf{r}^{\mathrm{1,2,\ldots,10}}_t
\right).
\end{equation}

Each privileged observation $\mathbf{o}^{\mathrm{priv}}_t\in\mathbb{R}^{459}$ contains the root position, linear velocity, orientation and angular velocity in the world frame, the body positions, orientations, linear and angular velocities in the robot heading frame, the joint positions and velocities, and the previous action. The critic is an MLP with four hidden layers of width 1024 and Exponential Linear Units (ELU)~\cite{clevert2015fast} as activations. It receives no $\theta$ features and is used only during training.

The actor is Transformer-based. At each control step, the actor receives
\begin{equation}
\mathbf{o}^{\mathrm{act}}_t
=\left[\mathbf{o}_{t-N+1:t};\,\mathbf{r}^{\mathrm{next}}_t;\,\mathbf{G}_t\right]
\in\mathbb{R}^{1182},
\end{equation}
where $N=10$ and $\mathbf{o}_t\in\mathbb{R}^{81}$ contains joint positions and velocities, the previous action, and heading-frame root angular velocity and orientation. The history contributes 810 features, $\mathbf{r}^{\mathrm{next}}_t\in\mathbb{R}^{52}$ specifies the next-step reference, and $\mathbf{G}_t\in\mathbb{R}^{320}$ is the reference generator's output, $\theta$ goal window that is explained in Sec. \ref{sec:motion_encoding}. The actor architecture is illustrated in Sec. \ref{sec:actor}.

\begin{table}[h]
\centering
\caption{Reward terms}
\label{tab:reward_terms}
\small
\setlength{\tabcolsep}{3pt}
\renewcommand{\arraystretch}{1.15}
\begin{tabular}{@{}l>{$}l<{$}r@{}}
\toprule
Term & \multicolumn{1}{l}{Definition} & Weight \\
\midrule
\multicolumn{3}{@{}l}{$r^{\mathrm{track}}$} \\
\quad Torso pos.     & \exp\bigl(-\lVert \hat{\mathbf{p}}^{A} - \mathbf{p}^{A} \rVert^2/\sigma_p^2\bigr) & $0.5$ \\
\quad Torso ori.     & \exp\bigl(-\angle(\hat{\mathbf{q}}^{A}, \mathbf{q}^{A})^2/\sigma_q^2\bigr) & $0.5$ \\
\quad Body pos.       & \exp\bigl(-\tfrac{1}{|\mathcal{B}|}\textstyle\sum_{b} \lVert \bar{\mathbf{p}}_b - \mathbf{p}_b \rVert^2/\sigma_p^2\bigr) & $1.0$ \\
\quad Body ori.       & \exp\bigl(-\tfrac{1}{|\mathcal{B}|}\textstyle\sum_{b} \angle(\bar{\mathbf{q}}_b, \mathbf{q}_b)^2/\sigma_q^2\bigr) & $1.0$ \\
\quad Body lin.\ vel. & \exp\bigl(-\tfrac{1}{|\mathcal{B}|}\textstyle\sum_{b} \lVert \hat{\mathbf{v}}_b - \mathbf{v}_b \rVert^2/\sigma_v^2\bigr) & $1.0$ \\
\quad Body ang.\ vel. & \exp\bigl(-\tfrac{1}{|\mathcal{B}|}\textstyle\sum_{b} \lVert \hat{\boldsymbol{\omega}}_b - \boldsymbol{\omega}_b \rVert^2/\sigma_\omega^2\bigr) & $1.0$ \\
\midrule
\multicolumn{3}{@{}l}{$r^{\mathrm{reg}}$} \\
\quad Action rate     & \lVert \mathbf{a}_t - \mathbf{a}_{t-1} \rVert^2 & $-0.3$ \\
\quad Joint limits    & \textstyle\sum_{j} \operatorname{dist}\bigl(\xi_j, [\underline{\xi}_j, \overline{\xi}_j]\bigr) & $-1.0$ \\
\quad Contacts        & \textstyle\sum_{b \notin \mathcal{B}_{\mathrm{allow}}} \mathbf{1}\bigl[\lVert \mathbf{F}_b \rVert > 1\,\mathrm{N}\bigr] & $-0.1$ \\
\bottomrule
\end{tabular}
\end{table}

\subsubsection{Whole-Body Tracking Reward}
\label{sec:reward}
We define the reward as the sum of a tracking part and a regularization part,
\begin{equation}
  r_t \;=\; r^{\mathrm{track}}\big(s_t, \hat s_t\big)
        \;+\; r^{\mathrm{reg}}\big(s_t, a_t, a_{t-1}\big),
  \label{eq:rew_total}
\end{equation}
where $r^{\mathrm{track}}$ minimizes body-link pose and
velocity errors against the reference state $\hat s_t$, and
$r^{\mathrm{reg}}$ penalizes jerky actions, joint-limit violations and
unintended contacts. Table~\ref{tab:reward_terms} lists the terms of both parts.

In Table~\ref{tab:reward_terms}, $\mathbf{p}$, $\mathbf{q}$, $\mathbf{v}$, $\boldsymbol{\omega}$ represent link position, orientation, linear and
angular velocity, respectively. Hatted variables denote the corresponding reference values. $\angle$ denotes the geodesic angular distance between the reference and actual orientations, measured in radians. $\mathcal{B}$ indicates the 12 tracked links ($|\mathcal{B}|=12$), and $\mathcal{B}_{\mathrm{allow}}$ denotes the links allowed to make contact, namely the feet and forearms ($|\mathcal{B}_{\mathrm{allow}}| = 4$). 

\subsubsection{Adaptive Sampling}
We combine randomized subdivision of long sequences with sampling weights based on completion history and tracking error. Weights decrease as motions are mastered, while poorly tracked segments receive higher priority. Normalizing these weights yields sampling probabilities that reduce repeated exposure to easy segments and emphasize motions requiring further improvement.

\subsection{Real-Time Monocular Video Processing Module (SEG)}
\subsubsection{Real-Time 3D Human Reconstruction from Video Streams}
\label{sec:Reconstruction}
Prior humanoid whole-body teleoperation systems either track only sparse head and hand targets or still depend on MoCap. We reconstruct human motion from a monocular RGB stream and transmit SMPL parameters to the robot retargeting module. Visual estimation runs asynchronously at approximately 10~Hz.

\textit{Per-Frame Monocular Reconstruction}: A TensorRT YOLO11-pose detector returns person boxes and 2D keypoints. Reconstruction proceeds only when exactly one detection exceeds a confidence threshold of 0.75.

SAM 3D Body processes the $512\times512$ person crop using a DINOv3~\cite{simeoni2025dinov3} backbone conditioned on camera rays derived from the intrinsics $K$. Intrinsics are obtained from fixed-camera calibration or per-frame MoGe-2 estimates~\cite{wang2025moge2accuratemonoculargeometry}. A promptable decoder and Momentum Human Rig (MHR)~\cite{ferguson2025mhr} head estimate mesh vertices $\mathbf{V}\in\mathbb{R}^{18439\times3}$, root translation $\mathbf{t}$, and global orientation, followed by inverse-kinematics refinement. Hand decoding is omitted from the streamed output.

\begin{figure}[h]
\includegraphics[width=\linewidth]{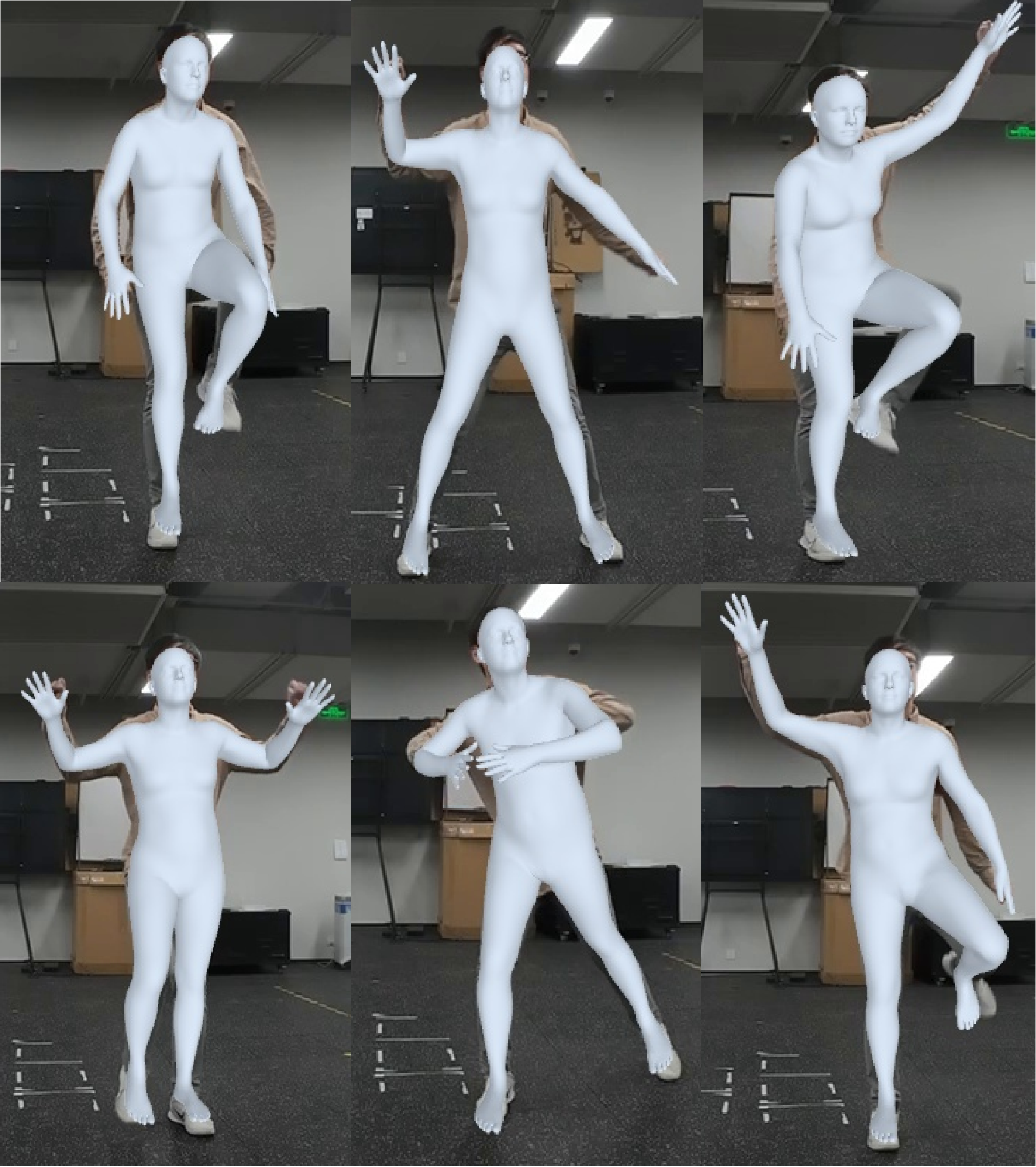}
    \centering
   \caption{Per-frame human motion reconstruction through the RGB-to-MHR-to-SMPL pipeline.}
   \label{SAM}
   %\vspace{-1.0em}
\label{fig:recon}
\end{figure}

The MHR mesh is mapped to Skinned Multi-Person Linear Model (SMPL)~\cite{SMPL-X:2019} by a precomputed barycentric correspondence:
SMPL vertex $j$ is $\hat{\mathbf{v}}_j=\sum_{k=1}^{3} b_{j,k}\,\mathbf{V}_{\mathrm{w}}[f_{j,k}]$, where $V_{\mathrm{w}}=V+\mathbf{t}$. A fixed subset of $1500$ vertices, where $\mathbf{V}_{\mathrm{w}}=\mathbf{V}+\mathbf{t}$. A fixed subset of $1500$ vertices, centroid-centered and flattened to $\mathbb{R}^{4500}$, is fed to a small MLP that regresses a 6D global orientation, the $63$-dimensional body pose
($21$ joints, axis-angle), and $10$ shape coefficients. SMPL forward kinematics yields $24$ joints, optionally denoised by a sliding-window smoothing MLP. The output results of the per-Frame monocular reconstruction module are shown in Fig.~\ref{fig:recon}.

For the fixed-camera setup, we use the calibrated gravity direction to express the reconstructed human motion in a gravity-aligned coordinate frame with an upward-pointing $z$-axis.

\subsubsection{Reference Motions Encoding and Frequency-Domain Representation}
\label{sec:motion_encoding}
Reference motions contain temporal structure that can inspire WBC tracking.
Most existing trackers treat the motion library as an unstructured collection of samples, so clips containing multiple behaviors collapse into a mixed descriptor. In contrast, we encode every short motion window, rather than an entire sequence, into per-channel sinusoidal latent parameters, namely a frequency, amplitude, and offset that describe the high-level features of the motion and a phase that indexes local time within it. Assuming that the frequency, amplitude, and offset remain quasi-constant over a short horizon, an encoder is trained with a multi-step forward-prediction loss, so that future states are predicted by advancing only the phase at the encoded frequency and decoding the result. These latent dynamics yield a compact, and temporally coherent representation. To reduce the impact of the local periodicity of this method and improve the tracking performance of less periodic motions such as dance, we remove the forward-prediction terms in order to train on local reconstruction alone. Its parameters are re-estimated from each new reference window rather than constrained to remain constant across a prediction horizon, so the frequency, amplitude, and offset can follow expressive, non-periodic changes in the motion. These two formulations thus serve complementary roles in MoViC. The local-reconstruction encoder provides an accurate, and time-varying description of the reference motion, which is used to construct the $\theta$ for policy conditioning and to characterize the reference motion library. Thanks to its motion prediction and genreation capacity, the predictive latent dynamics of the periodic representation are used during real-world deployment to generate reference motions that compensate the low-rate visual updates.

\textit{Motion Encoder:}
Each reference frame is a body state $\mathbf{s}_n\in\mathbb{R}^{57}$ comprising base-frame linear and angular velocities, projected gravity, and the positions and velocities of the 24 actuated joints. We construct a trailing window $\mathbf{S}_n=[\mathbf{s}_{n-H+1},\ldots,\mathbf{s}_n]\in\mathbb{R}^{57\times H}$ with $H=31$, spanning one second. A one-dimensional convolutional encoder maps the window to $c=64$ latent channels. At layer $\ell$, the temporal convolution is

\begin{equation}
U^{(\ell)}_{n,k,h}
=d^{(\ell)}_k+\sum_{j=1}^{C_{\ell-1}}\sum_{r=-15}^{15}
W^{(\ell)}_{k,j,r}\,\widetilde{X}^{(\ell-1)}_{n,j,h+r}.
\end{equation}

\noindent where $C_{\ell-1}$ is the input channel count, $W^{(\ell)}$ and $d^{(\ell)}$ are learned weights and biases, and $\widetilde{X}^{(\ell-1)}$ denotes the padded layer input, with $\mathbf{X}^{(0)}_n=\mathbf{S}_n$. The sums aggregate input channels $j$ and temporal offsets $r$ for output channel $k$ at position $h$. The kernel length is 31; unit stride and same padding preserve the window length, yielding encoder output $\mathbf{Z}_n\in\mathbb{R}^{c\times H}$.

A differentiable fast Fourier Transform (FFT) estimates the frequency $f_{n,k}$, amplitude $\alpha_{n,k}$, and offset $b_{n,k}$ of each channel $k$. A linear head followed by $\operatorname{atan2}$ estimates the phase $\phi_{n,k}$. Here, $\alpha$ denotes the Fourier amplitude to distinguish it from the policy action. The latent signal is reconstructed as

\begin{equation}
\widehat{z}_{n,k}(\tau)
=\alpha_{n,k}\sin\!\left(2\pi(f_{n,k}\tau+\phi_{n,k})\right)+b_{n,k},
\end{equation}

\noindent where $\tau$ is the temporal coordinate within the window. A decoder maps the reconstructed latent sequence to $\widehat{\mathbf{S}}_n$.

% We retain only the zero-step reconstruction term ($i=0$) of the FLD forecasting objective:
The periodic and local representations share the same encoder and decoder design, but differ in the training objective. Let $\boldsymbol{\eta}_n=(\mathbf{f}_n,\boldsymbol{\alpha}_n,\mathbf{b}_n)$ collect the frequencies, amplitudes, and offsets encoded from $\mathbf{S}_n$, and let $\mathcal{D}(\boldsymbol{\phi},\boldsymbol{\eta})=\operatorname{Dec}\big(\widehat{P}(\boldsymbol{\phi},\boldsymbol{\eta})\big)$, where $\widehat{P}$ evaluates the sinusoidal reconstruction of all latent channels. The two objectives are

\begin{subequations}\label{eq:fld_dfm}
\begin{align}
\mathcal{L}_{\mathrm{periodic}}
&=\sum_{i=0}^{M-1}\big\|
\mathcal{D}(\boldsymbol{\phi}_n+i\,\mathbf{f}_n\Delta t,\;\boldsymbol{\eta}_n)
-\mathbf{S}_{n+i}\big\|^{2},
\label{eq:fld_loss}\\
\mathcal{L}_{\mathrm{local}}
&=\big\|
\mathcal{D}(\boldsymbol{\phi}_n,\;\boldsymbol{\eta}_n)
-\mathbf{S}_{n}\big\|^{2},
\label{eq:dfm_reconstruction}
\end{align}
\end{subequations}
\noindent where $\Delta t$ is the reference frame interval and $M$ is the prediction horizon. In~\eqref{eq:fld_loss}, the parameters $\boldsymbol{\eta}_n$ encoded from one window must also reconstruct the following $M-1$ windows once the phase is advanced, which acts as the reference generator and enforces the latent dynamics used for extrapolation. Eq.~\eqref{eq:dfm_reconstruction} is the case $M=1$: each window is reconstructed only from its own encoding, so $\boldsymbol{\eta}_n$ can change from one window to the next. We train the encoder for the $\theta$ goal window used for actor, with~\eqref{eq:dfm_reconstruction}. This encoder acts as the reference motion encoder. Eq.~\eqref{eq:dfm_reconstruction} emphasizes local reconstruction without enforcing multi-step predictions from a fixed latent parameterization. The sinusoidal representation is retained, while its parameters are recomputed for each reference window to capture changes in the motion. 

\begin{table}[htbp]
\centering
\caption{Reference-Motion encoder training hyperparameters.}
\label{tab:dfm_hparams}
\small
\setlength{\tabcolsep}{4pt}
\renewcommand{\arraystretch}{1.05}
\begin{tabular}{@{}>{\raggedright\arraybackslash}p{0.30\columnwidth}
  >{\raggedright\arraybackslash}p{\dimexpr0.70\columnwidth-2\tabcolsep\relax}@{}}
\toprule
Parameter & Value \\
\midrule
Enc./dec. & Conv1D; hidden widths $[768,768,768]$ \\
State dimension & 57 per frame \\
Latent channels & $c=64$ \\
Kernel/padding & $31$/same \\
Normalization & Batch normalization (BN) \\ %~\cite{ioffe2015batch}
Activation & ELU \\
Loss & Mean Square Error \\
Spectral estimates & Real FFT: frequency $f_k$, amplitude $\alpha_k$, offset $b_k$ \\
$\theta$ output & $\big[\mathbf{f}_n;\boldsymbol{\alpha}_n;\mathbf{b}_n;$\newline
  $\sin(2\pi\boldsymbol{\phi}_n);\cos(2\pi\boldsymbol{\phi}_n)\big]$ \\
Input window & $H=31$ frames at $30$~Hz \\
% Train set & 61{,}702 clips \\
Batch size & 512 \\
\bottomrule
\end{tabular}
\par\smallskip
\begin{minipage}{\columnwidth}
\end{minipage}
\end{table}

After training, the encoder is frozen and applied at every reference frame. Windows extending before the start of a clip are padded by repeating its first frame. We store the phase-augmented representation

\begin{equation}
\begin{aligned}
\widetilde{\boldsymbol{\theta}}^{\mathrm{tail}}_n
&=\big[\mathbf{f}_n;\boldsymbol{\alpha}_n;\mathbf{b}_n;\\
&\qquad\sin(2\pi\boldsymbol{\phi}_n);\cos(2\pi\boldsymbol{\phi}_n)\big],\\
\widetilde{\boldsymbol{\theta}}^{\mathrm{tail}}_n
&\in\mathbb{R}^{5c},\qquad 5c=320.
\end{aligned}
\end{equation}

The sine--cosine encoding avoids discontinuities at phase wrapping. The five 64-dimensional parameter blocks yield a 320-dimensional reference representation.

\textit{Motion Representation:}
For motion library analysis, we average and standardize the reference motion encoder's 192-dimensional parameters $[\mathbf{f}_n;\boldsymbol{\alpha}_n;\mathbf{b}_n]$ over 3-s windows with a 1-s stride. Fifty principal components are then retained. These phase-free descriptors are used for analysis, while the actor receives the 320-dimensional phase-augmented codes.

\begin{figure}[h]
  \centering
  \includegraphics[width=\linewidth]{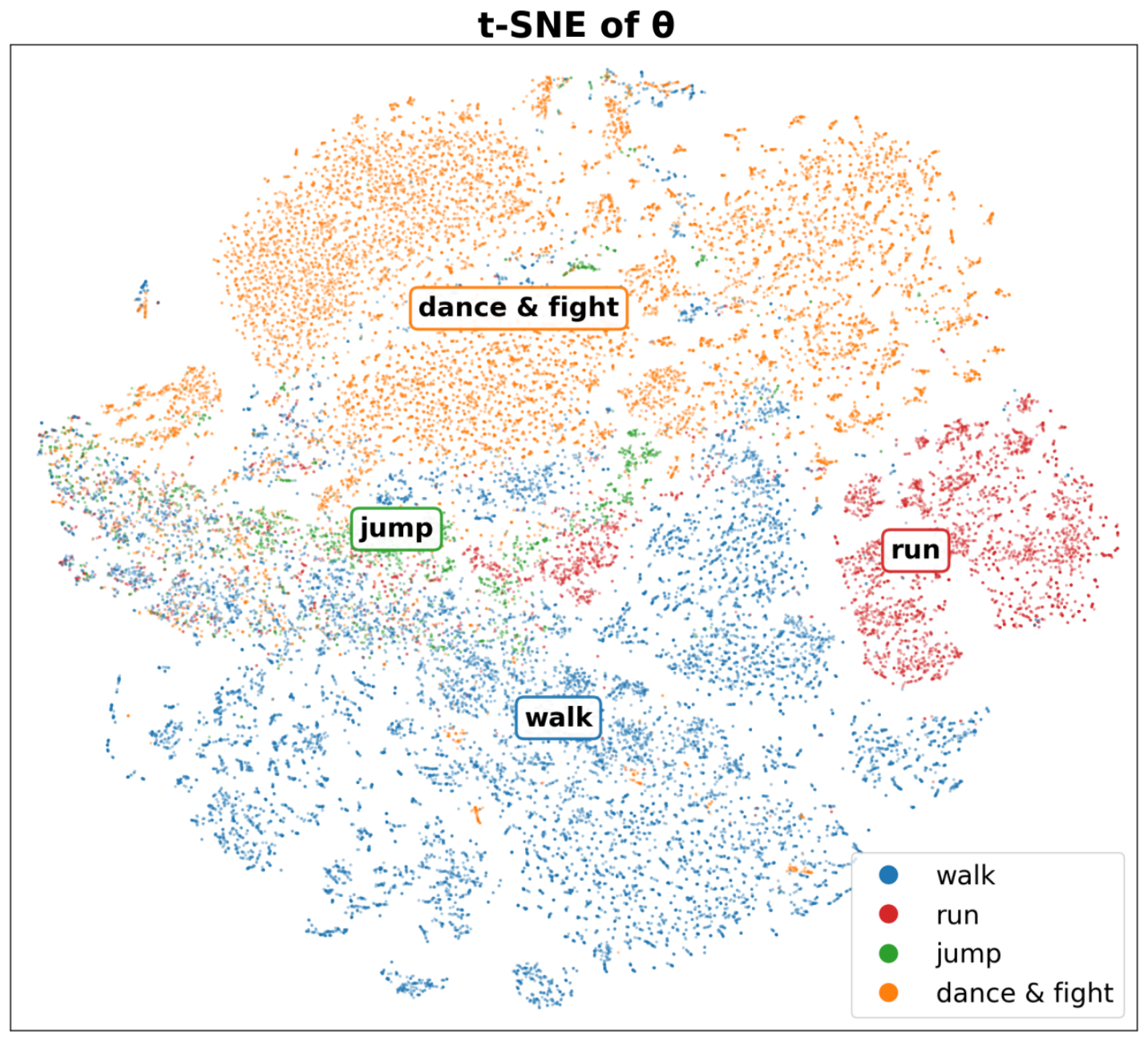}
  \caption{t-SNE visualization of DFM-derived motion descriptors. Each point represents a 3-s motion window, colored by category.}
  \label{fig:theta_tsne}
\end{figure}

We visualize these descriptors using t-SNE~\cite{vandermaaten2008tsne}. Fig.~\ref{fig:theta_tsne} shows partially separated motion categories with variation within categories and overlap between them; labels are used only for coloring.
% Full preprocessing and sampling details are retained in Section_III_B_refinement_notes.md.

$k$-NN ($k=15$) with clip-grouped five-fold cross-validation yields a classification accuracy of 0.91 per window and 0.94 after majority voting within each clip, compared with a chance level of approximately 0.42. On a common subset of 30{,}000 windows, clip-level accuracies are 0.72 for two principal components, 0.90 for two-dimensional UMAP, 0.93 for two-dimensional t-SNE, and 0.94 for the 50-dimensional PCA features. Preprocessing and embeddings are fitted on the full set, including windows later held out, before cross-validation; these results therefore describe transductive category discrimination.

Let $\Delta_c=1/50\,\mathrm{s}$ denote the control interval and $\tau_t=\tau_0+t\Delta_c$ the reference timestamp at control step $t$, where $\tau_0$ is the initial reference time. During training, the actor receives one explicit target $\mathbf{r}^{\mathrm{next}}_t$ ($K=1$) for time $\tau_t+\Delta_c$. Thus, the target corresponds to the next control step, rather than the next frame of the original reference stream.

The $\theta$ goal window provides additional temporal context. Training codes are indexed on the original 30-Hz reference grid. For a clip of length $T$, we define an offset code as

\begin{equation}
\widetilde{\boldsymbol{\theta}}^{(\delta)}_n
=\widetilde{\boldsymbol{\theta}}^{\mathrm{tail}}_
{\operatorname{clip}(n+\delta,\,0,\,T-1)},
\end{equation}

\noindent where $\delta$ is measured in reference frames. Let $n_t$ denote the reference-frame index corresponding to control time $\tau_t$. With $\delta=10$ and $H=31$, the future-aligned code summarizes $[n_t-20,n_t+10]$ away from clip boundaries, providing a ten-frame lookahead of approximately 0.333~s. The actor's complete goal window is denoted by

\begin{equation}
\mathbf{G}_t=\widetilde{\boldsymbol{\theta}}^{(+10)}_{n_t}\in\mathbb{R}^{320}.
\end{equation}

The explicit next-step target specifies the immediate tracking objective, while $\mathbf{G}_t$ summarizes additional reference context. 

\subsubsection{Latent-Dynamics Extrapolation of Low-Rate Visual References}
\label{sec:extrapolation}
\begin{figure}[h]
\includegraphics[width=0.5\textwidth]{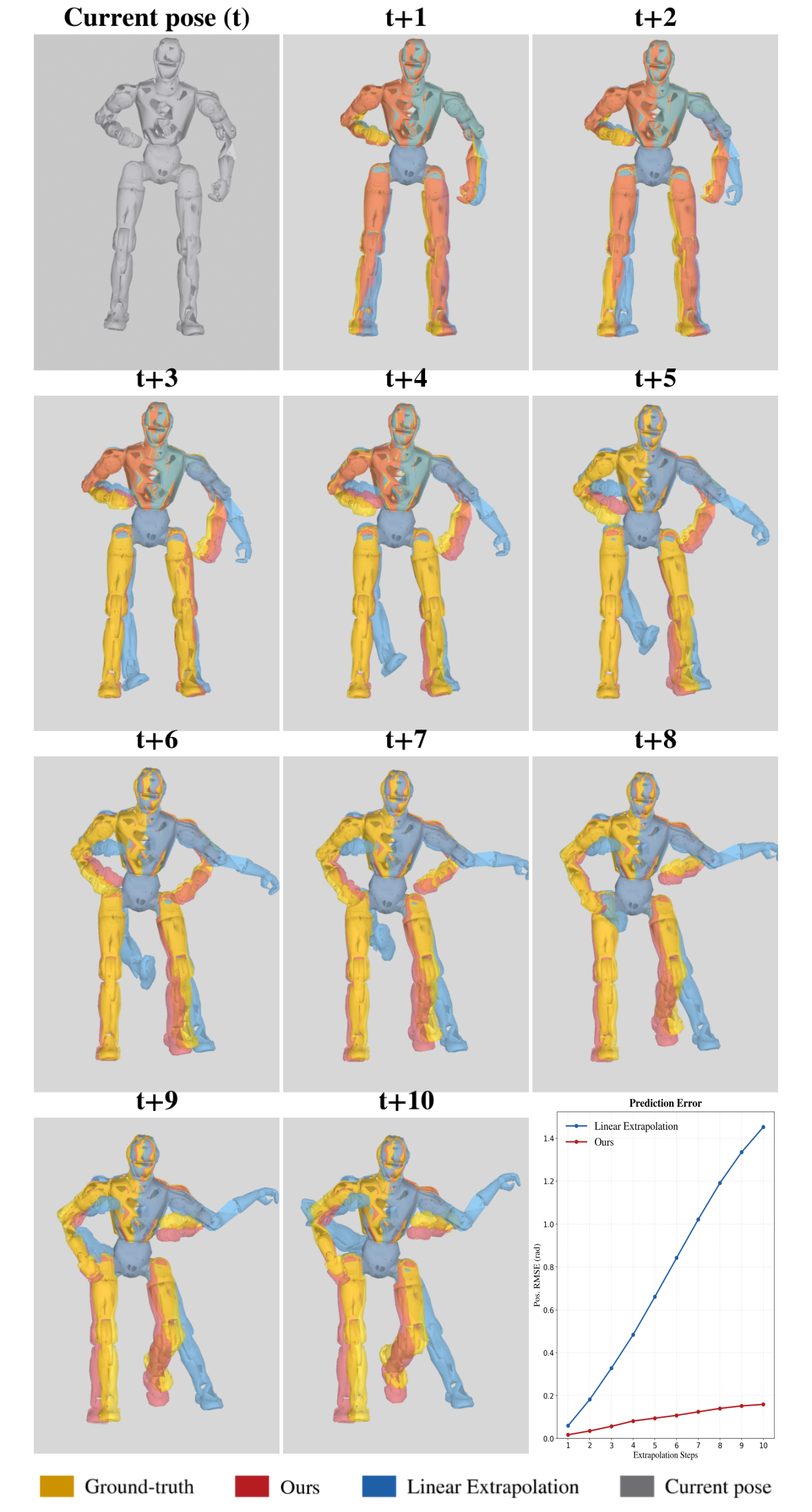}
    \centering
   \caption{\textbf{Our extrapolation method versus linear extrapolation}. The \textbf{current pose} is the robot state at the current time. Frames t+1 through t+10 overlay the ground-truth pose with FLD and linear extrapolation. In the prediction-error plot, error is the aggregate joint-position RMSE. The \textcolor{red}{red} line is \textcolor{red}{our} extrapolation method, and the \textcolor{blue}{blue} line is the \textcolor{blue}{linear} extrapolation. Linear-extrapolation error grows rapidly with the horizon and remains much larger than that of ours, so the our extrapolation method are substantially more accurate.}
   \label{sim:extrapolation}
   %\vspace{-1.0em}
\end{figure}
High-quality visual reconstruction runs much slower than the humanoid control loop, and this mismatch has so far been handled by reconstructing motion offline before execution or by leaving the tracker to tolerate the delayed, low-rate estimates. Rather than trading reconstruction quality for speed, we close this gap using the periodic latent dynamics extrapolation method. The reconstructed poses are retargeted to the robot and supplied to the reference generator in Sec.~\ref{sec:extrapolation}. This module fills the intervals between low-rate updates by advancing the latent phase and decoding reference states at 50~Hz. Predictions are anchored to the latest accepted reference using~\eqref{eq:anchored}.
At deployment, our extrapolation methods generates intermediate references for 50-Hz policy execution from an approximately 10-Hz motion stream. The extrapolation network is trained using largely identical hyperparameters to those in Table~\ref{tab:dfm_hparams}, except that it is trained with~\eqref{eq:fld_loss} and $M=15$ instead of~\eqref{eq:dfm_reconstruction}. Each accepted reference updates a rolling buffer of $H$ states, which is encoded into latent phase, frequency, amplitude, and offset. Between updates, the latent phase is advanced and the resulting states are decoded. This generates reference estimates without access to future measurements.

To anchor predictions to the latest measurement, we add the predicted deviation in the decoded motion to the observed state. Let $\mathbf{d}_{t+i}$ denote the decoded prediction $i$ steps ahead and $\mathbf{d}_t$ the reconstruction of the current frame:

\begin{equation}
  \hat{\mathbf{x}}_{t+i}
  \;=\;
  \mathbf{x}_t + \big(\mathbf{d}_{t+i} - \mathbf{d}_{t}\big),
  \qquad i = 1,\dots,k\textcolor{red}{.}
  \label{eq:anchored}
\end{equation}
Only the increment $\mathbf{d}_{t+i}-\mathbf{d}_t$ is added to the latest measurement $\mathbf{x}_t$. This cancels any reconstruction offset shared by the current and predicted decoded states, but does not eliminate differences between their reconstruction errors.

Incoming references are checked using the multi-step prediction error over the recent buffer. Updates exceeding a threshold calibrated on held-out data are rejected, and extrapolation continues from the last accepted state. This check measures consistency with the learned motion model; it does not guarantee that every erroneous visual estimate will be detected.

% \subsubsection{Extrapolation}
Table~\ref{tab:extrapolation} compares joint-position and joint-velocity RMSE for our method and linear extrapolation. Fig.~\ref{sim:extrapolation} illustrates the corresponding predicted configurations in simulation.

\begin{table}[t]
    \caption{Comparison of Different Extrapolation Methods}
    \label{tab:extrapolation}
    \centering
    \footnotesize
    \renewcommand{\arraystretch}{1.15}
    \setlength{\tabcolsep}{3pt}

    \begin{tabular}{crrrr}
        \toprule
        
        \multirow{2}{*}[-4pt]{}
        & \multicolumn{2}{c}{\textbf{FLD}}
        & \multicolumn{2}{c}{\textbf{Linear}} \\

        \cmidrule(lr){2-3}
        \cmidrule(lr){4-5}

        & \shortstack{\textbf{Pos. RMSE}\\(rad)}
        & \shortstack{\textbf{Vel. RMSE} \\(rad/s)}
        & \shortstack{\textbf{Pos. RMSE} \\(rad)}
        & \shortstack{\textbf{Vel. RMSE} \\(rad/s)} \\

        \midrule

        \textbf{Mean Error}
        & \textbf{0.0655}
        & \textbf{1.2372}
        & 0.7561
        & 8.6816 \\

        \bottomrule
    \end{tabular}

\end{table}

\subsection{Transformer Actor Architecture}
\label{sec:actor}
Existing universal trackers typically rely on teacher–student distillation or mixtures of experts to gain capacity, and on explicitly stacked future reference frames to obtain temporal context that inflates the policy input. Our actor is a single transformer conditioned on a compact $\theta$ goal window, which summarizes the upcoming reference motion in the frequency domain. The actor receives an 810-dimensional proprioceptive history, a 52-dimensional next-step reference, and a 320-dimensional $\theta$ goal window, for a total of 1182 input features. Input streams are normalized and embedded by MLPs into tokens of dimension $D=768$. A learnable query provides the policy readout. Let $\mathbf{G}_{t,j}$ denote the blocks of the $\theta$ goal window:

\begin{equation}
\begin{aligned}
\mathbf{u}^{q}&=\mathbf{q}_{\mathrm{qry}},\\
\mathbf{u}^{\mathrm{prop}}_t
&=\operatorname{MLP}_{p}(\mathbf{o}_{t-N+1:t}),\\
\mathbf{u}^{\mathrm{ref}}_t
&=\operatorname{MLP}_{r}(\mathbf{r}^{\mathrm{next}}_t),\\
\mathbf{u}^{\theta}_{t,j}
&=\operatorname{MLP}_{\theta}(\mathbf{G}_{t,j}),
\quad j\in\{1,2\},
\end{aligned}
\end{equation}

\noindent where $N=10$, and $\mathbf{G}_{t,j}$ is the $j$-th block of $\mathbf{G}_t$. $\operatorname{MLP}_{p}$, $\operatorname{MLP}_{r}$, and $\operatorname{MLP}_{\theta}$ embed the proprioceptive history, the next-step reference, and the frequency-domain feature blocks, respectively, into $D$-dimensional tokens. These three networks have separate parameters, while $\operatorname{MLP}_{\theta}$ is shared across the two frequency-domain tokens. The query $\mathbf{q}_{\mathrm{qry}}\in\mathbb{R}^{D}$ is a learned parameter shared across control steps and environments; unlike the other tokens, it is not initialized from an observation or reference feature.

Each proprioceptive observation $\mathbf{o}_t\in\mathbb{R}^{81}$ contains joint positions and velocities, the previous action, and heading-frame root angular velocity and orientation. Orientation uses a six-dimensional tangent--normal representation. The next-step reference $\mathbf{r}^{\mathrm{next}}_t\in\mathbb{R}^{52}$ specifies the explicit tracking target at the next control timestamp. The $\theta$ features convey temporal context without explicitly stacking ten future reference frames in the proposed actor input.

Learned token-type embeddings and sinusoidal positional encodings are added to the sequence

\begin{equation}
\mathbf{X}_t=\left[
\mathbf{u}^{q},
\mathbf{u}^{\mathrm{prop}}_t,
\mathbf{u}^{\mathrm{ref}}_t,
\mathbf{u}^{\theta}_{t,1},
\mathbf{u}^{\theta}_{t,2}
\right].
\end{equation}

The query, proprioceptive input, reference target, and frequency-domain inputs use four distinct token types. The two frequency-domain tokens share a type embedding and are distinguished by position. The sequence is processed by four post-normalization transformer encoder layers with full bidirectional self-attention. Each layer uses six attention heads of dimension 128 and a feed-forward network of width 1024 with ReLU activations and no dropout.

The encoder output at the query position is normalized and passed to the action head:

\begin{equation}
\begin{aligned}
\mathbf{y}_t
&=\operatorname{LN}\!\left([\operatorname{Enc}_{T}(\mathbf{X}_t)]_0\right),\\
\boldsymbol{\mu}_t
&=\tanh\!\left(\operatorname{MLP}_{\mathrm{head}}(\mathbf{y}_t)\right),\\
\mathbf{a}_t
&\sim\mathcal{N}\!\left(\boldsymbol{\mu}_t,\sigma_a^2 I_{24}\right).
\end{aligned}
\end{equation}

The action head has three hidden layers of width 1024 with ELU activations. 

\textit{Training Protocol:} The controller is trained from scratch using proximal policy optimization (PPO), while the reference motion encoder pretrained with Eq.~\eqref{eq:dfm_reconstruction} remains fixed. Let $w$ and $\psi$ denote the actor and critic parameters, respectively. For each rollout batch, the principal actor--critic objective is:
\begin{equation}
\begin{gathered}
\underset{w,\psi}{\operatorname{arg\,min}}\Bigl\{-\mathcal{L}_{\mathrm{PPO}}^{\mathrm{clip}}(w;\mathbf{z}_t)+\lambda_V\mathcal{L}_{V}^{\mathrm{clip}}(\psi)\Bigr\},
\\[3pt]
\mathbf{z}_t=
\bigl[\mathbf{o}_{t-9:t};\mathbf{r}^{\mathrm{next}}_t;
\operatorname{sg}[\mathbf{G}_t]\bigr].
\end{gathered}
\label{eq:ppo_frozen_reference}
\end{equation}
where $\mathcal{L}_{\mathrm{PPO}}^{\mathrm{clip}}$ is the clipped policy surrogate evaluated over the rollout batch, $\mathcal{L}_{V}^{\mathrm{clip}}$ is the clipped value loss, and $\lambda_V$ weights the value loss. The stop-gradient operator $\operatorname{sg}$ preserves its input in the forward pass and blocks gradients in backpropagation.

We use a clipping threshold of $\epsilon=0.1$ and generalized advantage estimation with $\gamma=0.99$ and $\lambda=0.95$. Advantages and value targets are normalized. The Training is conducted on four A100 GPUs.

\section{Experiments and Discussion}
\subsection{Implementation Details}
Live video input is provided by a Logitech BRIO Ultra HD webcam, which is rigidly mounted and kept level. We capture $1920\times1080$ at $30$\,fps. Since the mount is level, the gravity direction is fixed to $(0,1,0)$ in the camera frame. The camera intrinsics are calibrated using the MoGe-2 FOV estimator. All models run on a PC with one NVIDIA RTX~5090.

\begin{figure*}[t]
\includegraphics[width=0.95\textwidth]{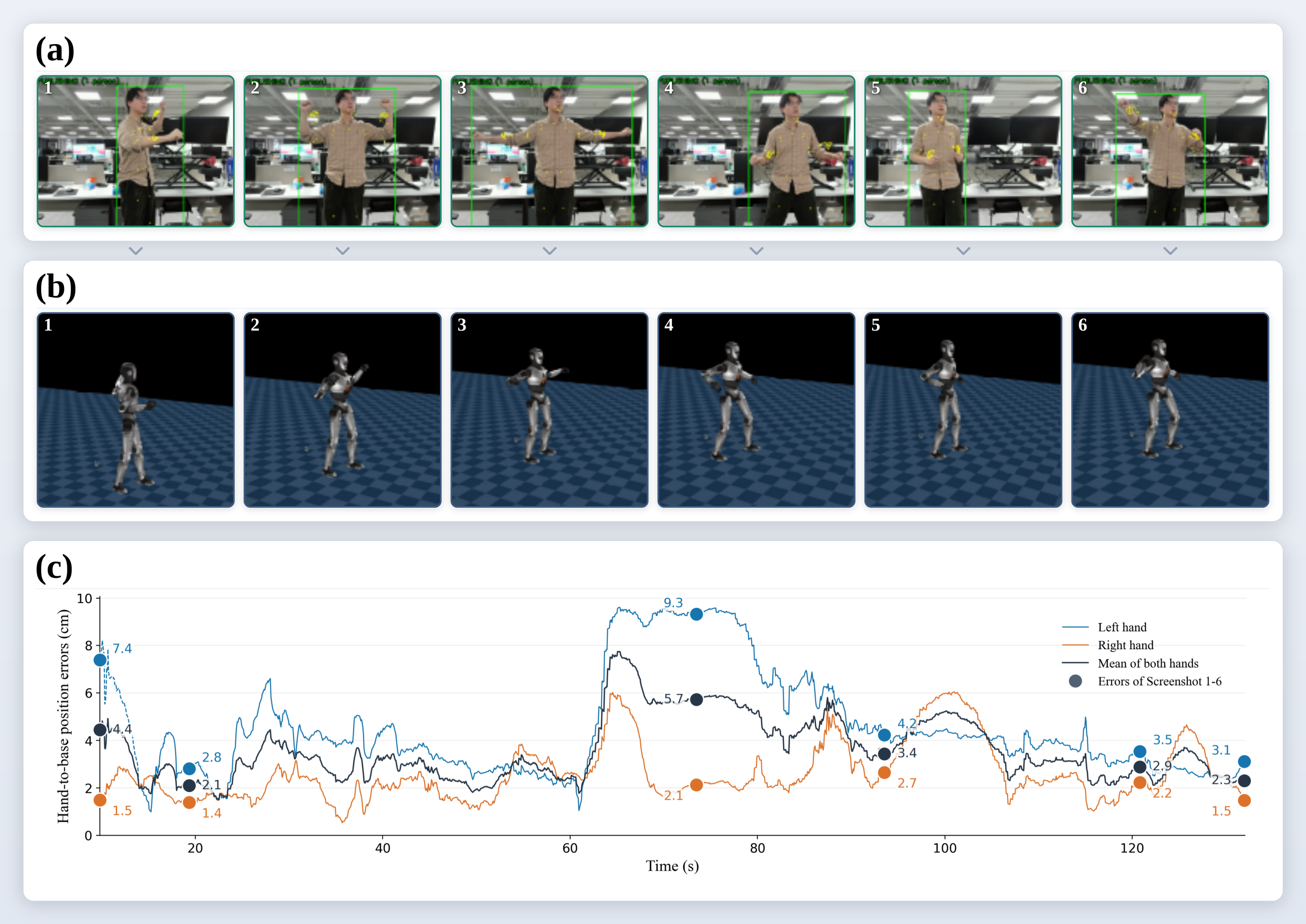}
    \centering
   \caption{\textbf{Simulation}. The simulation experiments are conducted in the MuJoCo simulator. \textbf{(a)} presents representative frames of the human motion together with the estimated body pose. Camera frames of the operator, showing the detected person (green bounding box) and the estimated body (yellow). \textbf{(b)} shows the corresponding controlled humanoid in simulation. Across different stages of the motion, the simulated robot closely follows the temporal evolution of the reference. \textbf{(c)} represents the hand-to-base position errors of the left hand (blue), right hand (orange) and their mean (dark). The markers on the plots indicate the errors at the instants of screenshots 1–6, corresponding in order to the frames shown in (a) and (b). At each instant, the blue, orange and dark markers give the left-hand, right-hand and mean errors, respectively}
   \label{sim_exp}
   %\vspace{-1.0em}
\end{figure*}

During deployment, capture, inference, and publication run asynchronously. A single-slot image buffer retains the latest timestamped frame for reconstruction (Section~\ref{sec:Reconstruction}). Reconstructed poses are retargeted, and the extrapolation supplies references at 50~Hz.

\subsection{Simulation}
We evaluate the complete MoViC pipeline in MuJoCo using RGB observations of a human performer as input. This end-to-end evaluation integrates visual motion reconstruction, motion retargeting, reference-rate adaptation, and policy execution, testing the full video-to-robot system rather than evaluating the tracking controller in isolation on preprocessed reference trajectories.

The results in Fig.~\ref{sim_exp} qualitatively illustrate consistency between the characteristic motion configurations of the human performer and those of the simulated robot, demonstrating the operation of the integrated pipeline from visual input to robot execution. The tracking errors of two end-effectors (right hand and left hand) are chosen to illustrate the tracking performance. The error are computed as Eq.\ref{eq:error}.
\begin{equation}
\label{eq:error}
e(t)=\left\|
\mathbf{r}_{\mathrm{GMR}}(t)-\mathbf{r}_{\mathrm{video}}(t)
\right\|_2.
\end{equation}

\noindent where $\mathbf{r}_{\mathrm{GMR}}(t)$ and
$\mathbf{r}_{\mathrm{video}}(t)$ denote the hand positions relative to
the base of the GMR reference robot and the robot reconstructed from
the video, respectively, each expressed in its own base coordinate frame. This evaluation complements the subsequent physical deployment experiments, while quantitative comparisons of tracking errors are presented in Sec. \ref{sec:comp}.

\subsection{Ablation Study}
In order to demonstrate the effectiveness of the $\theta$ goal window, ablation experiments on the policy observation are designed. We investigate whether the frequency-domain representation can replace an explicit multi-frame reference preview for whole-body motion tracking. All five variants receive an 810-dimensional proprioceptive history comprising ten consecutive 81-dimensional observations. Each explicit reference frame contains 52 features, and the $\theta$ goal window input contributes 320 features. The proposed configuration, $\theta$ with 1 frame, combines the proprioceptive history with the explicit next-step reference and a $\theta$ goal window containing information about ten future reference frames. The $\theta$ with 10 frames variant additionally provides nine explicit future reference frames. The 1 future frame and 10 future frames variants omit the $\theta$ goal window and use only the proprioceptive history and the corresponding explicit reference inputs. 

\begin{table}[htbp]
    \centering
    \caption{Ablation results on policy input.}
    \label{tab:ablation_final}
    \small
    \setlength{\tabcolsep}{4pt}
    \resizebox{\columnwidth}{!}{%
    \begin{tabular}{lcccc}
        \toprule
        Method & \shortstack{State Dimension\\$\mathcal{S}$} & \shortstack{MPJPE\\(rad)} & \shortstack{MPJVE\\(rad/s)} & \shortstack{Success rate\\(\%)} \\
        \midrule
        % CNN Enc with 10 frames & $\mathcal{S}\in\mathbb{R}^{1330}$ & 0.54 & 4.65 & 72\\
        1 future frame & $\mathcal{S}\in\mathbb{R}^{862}$ & 0.47 & 4.80 & 90 \\
        10 future frames & $\mathcal{S}\in\mathbb{R}^{1330}$ & 0.28 & 4.55 & 95  \\
        $\theta$ with 10 frames & $\mathcal{S}\in\mathbb{R}^{1650}$ & 0.28 & 4.45 & 99 \\
        $\theta$ with 1 frame & $\mathcal{S}\in\mathbb{R}^{1182}$ & \textbf{0.25} & \textbf{4.30} & \textbf{99} \\
        \bottomrule
    \end{tabular}
    }
\end{table}

Table~\ref{tab:ablation_final} shows that temporal reference context improves tracking performance. Without $\theta$, increasing the explicit preview from one to ten frames raises the success rate from 90\% to 95\%, while reducing MPJPE from 0.47 to 0.28~rad and MPJVE from 4.80 to 4.55~rad/s. The proposed $\theta$ with 1 frame configuration matches the highest success rate of 99\% and achieves the lowest MPJPE and MPJVE, 0.25~rad and 4.30~rad/s, respectively. Compared with 10 future frames, it improves the success rate by 2 percentage points and reduces MPJPE and MPJVE by 10.7\% and 5.5\%, respectively. These results support using $\theta$ to convey the additional reference context while retaining only the next-step target explicitly.

\begin{table*}[b]%[!t]
    \centering
    \caption{Comparison of Real-time Video Reconstruction Module}
    \label{tab:video_comp}
    \footnotesize
    \renewcommand{\arraystretch}{1.2}
    \setlength{\tabcolsep}{2.5pt}

    \begin{tabular*}{\textwidth}{@{\extracolsep{\fill}}l|*{5}{c}|*{5}{c}@{}}
        \toprule
        & \multicolumn{5}{c|}{\textbf{EMDB~\cite{kaufmann2023emdb}}}
        & \multicolumn{5}{c}{\textbf{Bencaogangmu}} \\
        \cmidrule(lr){2-6} \cmidrule(lr){7-11}
        \textbf{Methods} & \shortstack{MPJPE $\downarrow$\\(mm)}
        & \shortstack{PA-MPJPE $\downarrow$\\(mm)}
        & \shortstack{PVE $\downarrow$\\(mm)}
        & \shortstack{3D PCK $\uparrow$\\(\%)}
        & \shortstack{AUC $\uparrow$\\(\%)}
        & \shortstack{MPJPE $\downarrow$\\(mm)}
        & \shortstack{PA-MPJPE $\downarrow$\\(mm)}
        & \shortstack{PVE $\downarrow$\\(mm)}
        & \shortstack{3D PCK $\uparrow$\\(\%)}
        & \shortstack{AUC $\uparrow$\\(\%)} \\
        \midrule
        GV.+G & 123.45 & 75.15 & 135.3 & 58.14 & 38.82 & 126.5 & 102 & 117.25 & 52.15 & \textbf{31.81} \\
        Hy.+G & 189 & 122.76 & 225 & 47.92 & 27.15 & 150.875 & 104.375 & 144.25 & 45.83 & 27.36 \\
        SEG & \textbf{117.54} & \textbf{68.76} & \textbf{130.5} & \textbf{60.14} & \textbf{45.49} & \textbf{123.625} & \textbf{96} & \textbf{114.5} & \textbf{53.19} & 31.46 \\
        \bottomrule
    \end{tabular*}
    \par\vspace{3pt}
    \parbox{\textwidth}{\footnotesize
        GV.+G: GVHMR + GMR; Hy.+G: HybrIK + GMR; PCK threshold: 150\,mm; AUC range: 0--150\,mm. $\uparrow$: higher is better; $\downarrow$: lower is better.}
\end{table*}

The 90\% success rate of the one-frame baseline indicates that the immediate target already supports successful tracking for much of the evaluated motion set. Additional context is therefore motivated by the remaining failures, particularly segments that may require anticipation beyond the next target. The increase from 90\% to 95\% with ten explicit frames represents five percentage points of additional success, but an explicit ten-frame preview incurs a larger input at every control step, including portions where local guidance is sufficient. This motivates encoding extended reference context compactly. The results show that $\theta$ reaches 99\% success with one explicit target, exceeding the 95\% obtained from ten explicit frames. The aggregate success rates support this trade-off, although they do not identify which individual trajectories account for the improvement.

The benefit of $\theta$ is also observed when ten explicit reference frames are retained. This configuration does not outperform $\theta$ with 1 frame on any reported metric: both achieve 99\% success, while the single-frame configuration has lower position and velocity errors. Thus, providing nine additional explicit reference frames yields no observed improvement once the actor receives the frequency-domain representation, supporting the use of a single explicit next-step target alongside $\theta$.

The proposed configuration also uses a more compact actor input. Its input dimension is 1182, compared with 1330 for 10 future frames and 1650 for $\theta$ with 10 frames, corresponding to reductions of 11.1\% and 28.4\%, respectively. The reference component contains 372 features, comprising the 320-dimensional $\theta$ code and the 52-dimensional next-step target, compared with 520 features for ten explicit frames. Relative to $\theta$ with 10 frames, reducing the explicit reference input from ten frames to one also decreases the number of weights in the reference embedding's input layer when its hidden widths are held fixed. For a first-layer width of $d$, this removes $468d$ input weights. Smaller observation vectors also reduce rollout-storage requirements. The reduced input dimensionality therefore, accompanies equal or higher tracking success and lower tracking errors in these comparisons.

\subsection{Comparison Study}
\label{sec:comp}
To demonstrate our framework's ability in solving the tracking problems with the monocular videos, we design two comparison experiments to evaluate the quality of robot motions reconstructed from video in real time and the tracking performance of the overall framework, respectively.

\subsubsection{Comparisons of Real-time Video Reconstruction}
Table~\ref{tab:video_comp} evaluates the real-time video reconstruction module on EMDB dataset~\cite{kaufmann2023emdb} and Bencaogangmu. Following WHAM~\cite{shin2024wham}, and GVHMR~\cite{shen2026gvhmr}, we use the EMDB-1 split and evaluate only the frames marked valid by the official good-frame mask (17 sequences, 13.5 minutes). Bencaogangmu is a highly dynamic dance with 1400 frames in 50~Hz. We use the MoCap system to get the human motion data of Bencaogangmu and retarget it to the Engineai PM01 humanoids. The retargeted motion data acts as the ground truth for the comparison.  on the EMDB dataset and Bencaogangmu, the video reconstruction module of our framework, SEG that is composed of SAM 3D Body and FLD extrapolation, is compared with the HybrIK~\cite{li2021hybrik} used in OmniH2O~\cite{he2024omnih2o}, and the method used in SONIC~\cite{luo2026sonic} (GVHMR + sliding windows). To illustrate the real-time reconstruction performance, the output motion are of 50~Hz. Since GVHMR needs the video sequence as input and cannot run stably in 50~Hz, we adopt the linear exploration to compensate the gaps.

\begin{figure*}[t]
\includegraphics[width=0.9\textwidth]{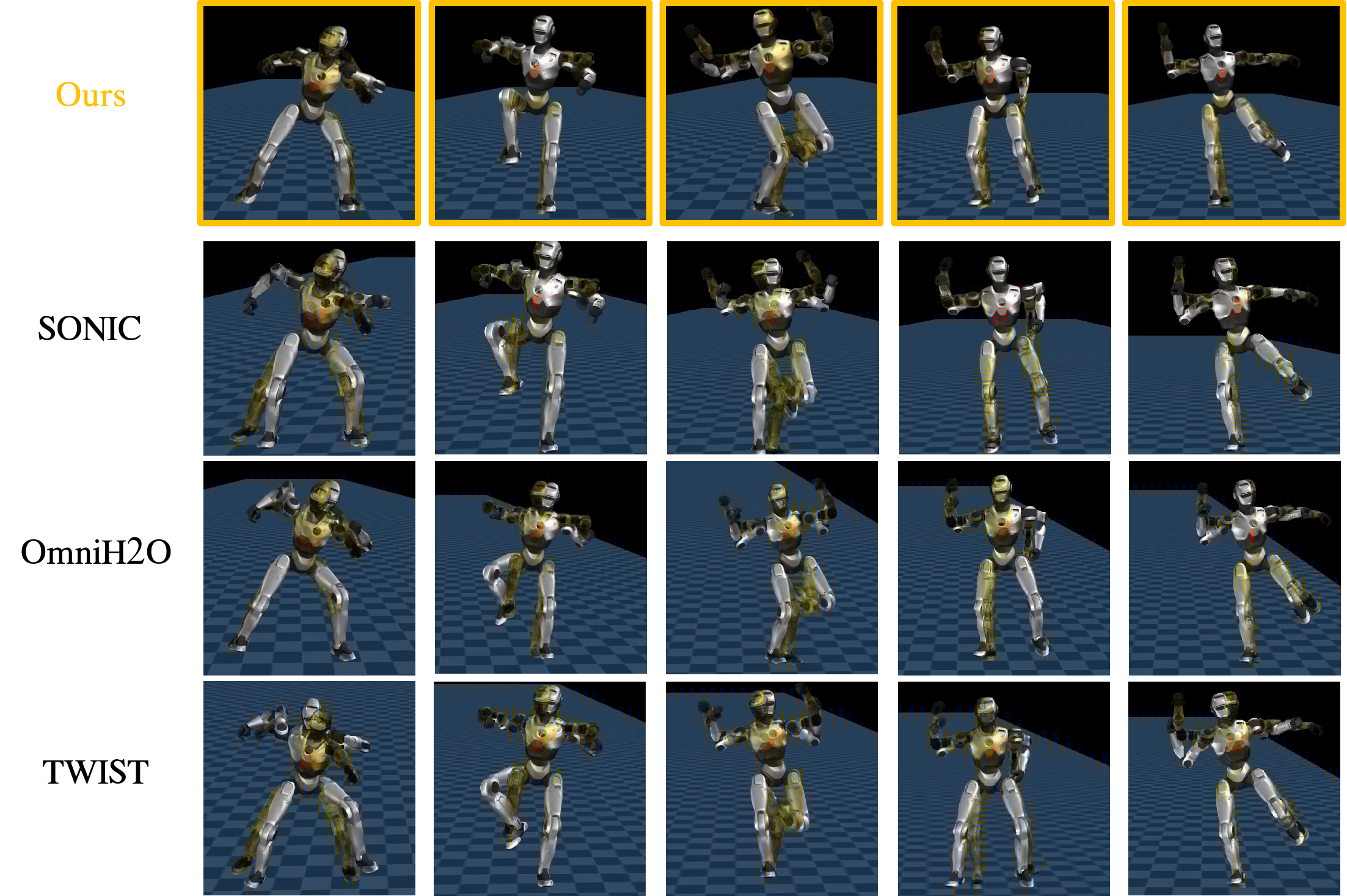}
    \centering
   \caption{\textbf{Simulation results of different methods}. In each simulation experiment, we visualize both the robot model and the input reference motion, where the latter is represented by a yellow ghost robot. The reference motion is reconstructed from a dance video (Bencaogangmu). The tracking performance of each method can be judged by how closely the controlled robot aligns with the ghost robot.}
   \label{pics:comaprison_sim_exp}
\end{figure*}

Following the evaluation protocol of WHAM~\cite{shin2024wham}, SLAHMR~\cite{ye2023slahmr}, and GVHMR~\cite{shen2026gvhmr}, we report the standard MPJPE, Procrustes-aligned MPJPE (PA-MPJPE) and per-vertex error (PVE) in mm to measure 3D pose accuracy, 3D PCK (150\,mm) and AUC~\cite{mono-3dhp2017} as threshold-based 3D accuracy, and the acceleration error (Accel, in m/s$^2$)~\cite{shin2024wham} to measure temporal smoothness, as shown in Table~\ref{tab:video_comp}.

According to Table~\ref{tab:video_comp}, SEG achieves the best results on all five metrics on EMDB and four of the five metrics on Bencaogangmu. On EMDB, SEG reduces MPJPE from 123.45 to 117.54 mm compared with GV. + G, while increasing 3D PCK and AUC by 2.00 and 6.67 percentage points, respectively. The relative reduction is largest for PA-MPJPE (8.5\%, compared with 4.8\% for MPJPE), indicating that the improvement mainly comes from more accurate local joint articulation rather than global orientation or scale. Moreover, the gain in AUC is considerably larger than that in 3D PCK, suggesting that SEG places more joints within tight error thresholds, i.e., it yields finer-grained pose accuracy rather than merely fewer gross failures. On Bencaogangmu, SEG achieves the lowest MPJPE, PA-MPJPE, and PVE, together with the highest 3D PCK of 53.19\%. Since SEG attains a higher 3D PCK at 150 mm but a slightly lower AUC averaged over 0–150 mm, it reduces large errors near the threshold. In addition, all methods show higher PA-MPJPE on Bencaogangmu (96.0–104.4 mm) than on EMDB, and the gaps between them are much smaller, suggesting that the various highly dynamic motions in Bencaogangmu makes this dataset challenging for all methods.

\subsubsection{Comparisons of tracking ability}
We compare our framework with the following three state-of-the-art trackers that support vision-based tracking - SONIC~\cite{luo2026sonic}, OmniH2O~\cite{he2024omnih2o}, and TWIST~\cite{ze2025twist}. To compare these frameworks on vision-based tracking, we design a comparison experiment using the same unseen dancing motion (Bencaogangmu). We use three metrics to evaluate the imitation error.
\begin{itemize}
    \item Mean per-joint position error (MPJPE): The absolute joint-angle error $\lvert q_{sim}-q_{ref}\rvert$ is computed over 23 joints, first averaged across joints and then across all frames in the sequence.
    \item Mean per-joint velocity error (MPJVE): Computed in the same manner as MPJPE, but using joint angular velocities instead of joint angles.
    \item Mean per-keypoint position error (MPKPE): The mean Euclidean distance between the 14 tracked keypoints selected from the robot.
\end{itemize}

\begin{figure}[h]
\includegraphics[width=0.45\textwidth]{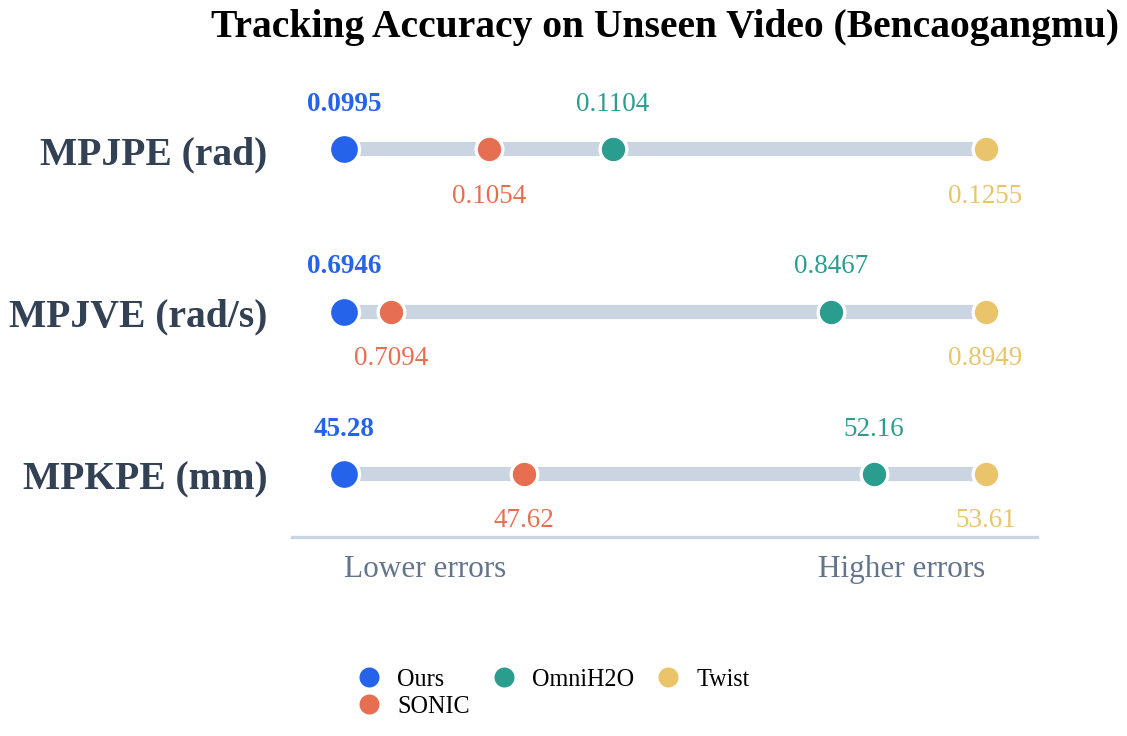}
    \centering
   \caption{Tracking accuracy on the unseen Bencaogangmu video. Lower is better.}
   \label{fig:comparison_data}
\end{figure}

All three error metrics above are calculated using the robot motions reconstructed from the video that act as the input reference motions of each tracker, as the ground truth. For a fair comparison, we reproduce all three baseline methods using their publicly released codebases and adapt them to the EngineAI PM01 humanoid robot. Fig. \ref{pics:comaprison_sim_exp} showcases the tracking performance from video input of different methods. Each baseline is trained on the same reference motion dataset used to train our method, ensuring that all methods are evaluated under a consistent embodiment. As TWIST does not have original video reconstruction module, we combine our reconstruction module with it.

All evaluations are conducted in the MuJoCo simulator, and the quantitative results are reported in Fig. \ref{fig:comparison_data}. Qualitative results illustrate that MoViC is able to reproduce more accurate and stable actions when tracking highly dynamic motions from videos. Improvements can be observed across all three metrics. MoViC obtains the lowest error on all these metrics. 

\subsection{Real-robot Experiments}
\begin{figure}[h]
\includegraphics[width=0.48\textwidth]{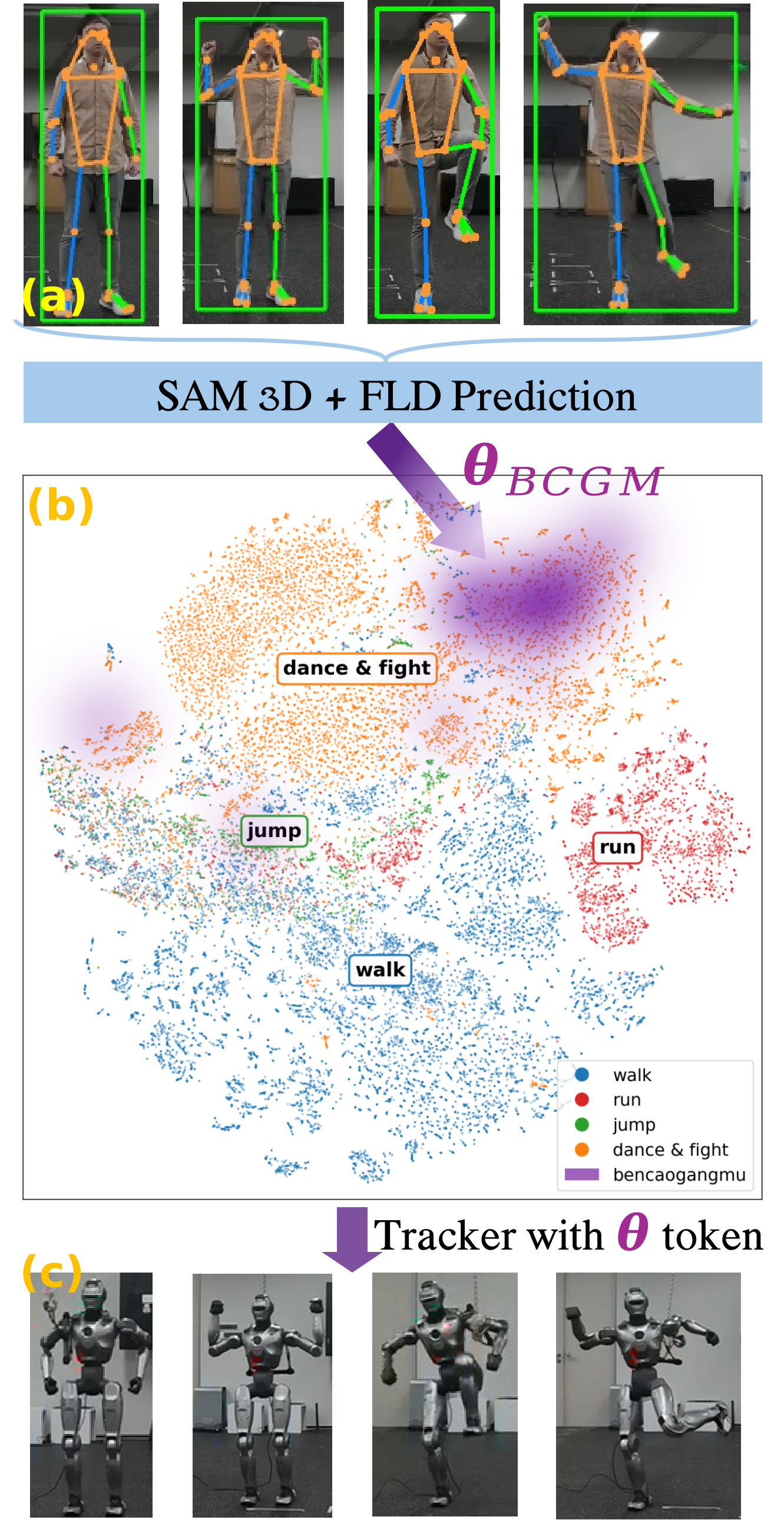}
    \centering
   \caption{\textbf{Real-world experiment}. (a)SAM 3D Body reconstructs the performer from monocular RGB. (b)The purple-highlighted BCGM region overlaps with the dance-and-fight region of the library. This overlap suggests category-level consistency between the video-derived dance representation and the library's dance-related representations in the visualization, connecting the deployment example to the motion-distribution analysis. (c)The physical robot executes the corresponding whole-body motion.}
   \label{fig:real_exp}
   %\vspace{-1.0em}
\end{figure}

\begin{figure*}[t]
\includegraphics[width=0.95\textwidth]{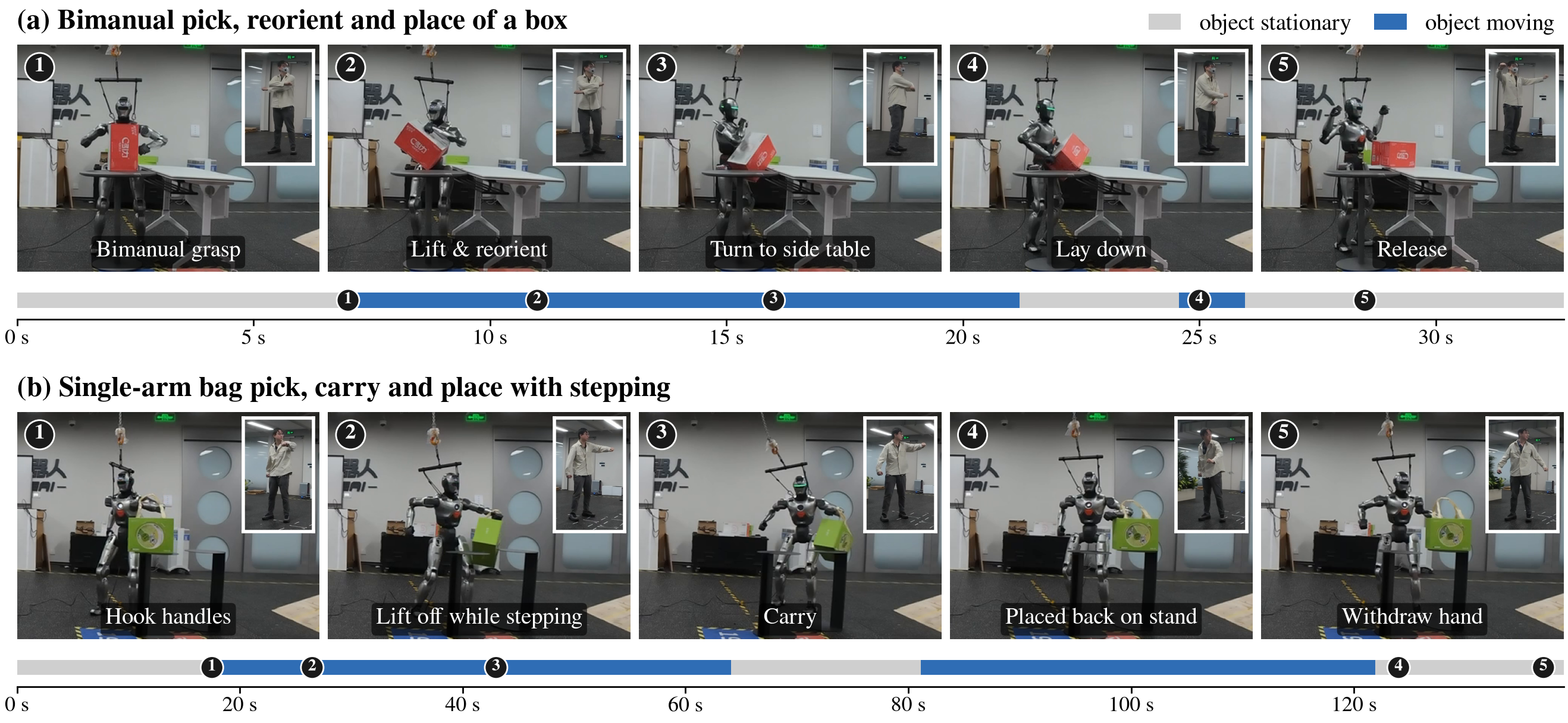}
    \centering
   \caption{\textbf{Teleoperation}. (a) Bi-manual pick, reorient, and place. (b) Single-arm pick, carry, and place in a 139-s continuous session. Insets show the operator, who wears no tracking device, as recorded by the third-person camera rather than the MoViC input camera.}
   \label{fig:teleop_manip}
\end{figure*}

In order to further demonstrate the practical effectiveness of the proposed framework, we conduct highly dynamic real-world experiments on the physical PM01 platform. These experiments are designed to evaluate whether the learned controller can retain its motion-tracking capability beyond simulation and execute the reference behaviors reliably under real robot dynamics and sensing conditions. The results are shown in Fig.~\ref{fig:real_exp} and~\ref{fig:teleop_manip}.

\textit{Tracking of Highly Dynamic Motions:} We use Bencaogangmu (BCGM), a well-known dance, for the real-world deployment test. The RGB signal processing module first detects and extracts the human pose at 10~Hz using SAM 3D Body. GMR~\cite{joao2025gmr} maps the human motion to the robot configuration in real time. FLD then extrapolates the retargeted motion from 10~Hz to 50~Hz, which aligns the reference rate with the control rate. The performance of the RGB signal processing module is shown in Fig.~\ref{fig:real_exp}-(a). 

To relate the deployment motion to the structure of the reference motion library, we also record the $\theta$ values of this video-reconstructed motion sequence. The distribution of $\theta_{BCGM}$ is showcased in Fig. \ref{fig:real_exp}-(b). It provides a qualitative view of this relationship rather than a quantitative measure of distribution alignment or tracking accuracy. 

The $\theta$ values play an important role in the motion tracker. Fig.~\ref{fig:real_exp}-(c) shows representative poses executed by the physical robot under the actions of our transformer-based BFM. These behaviors illustrate the reproduction of characteristic upper- and lower-body configurations of the dance, including single-leg-support poses with the other leg lifted. Together, the three panels provide a qualitative demonstration of the complete video-to-robot pipeline, linking visual motion estimates and their latent representations to coordinated whole-body execution on the physical platform.

\textit{Teleoperation through Vision:} Beyond free-space motion imitation, we evaluate MoViC as teleoperation tasks for object manipulation, in which the robot must establish physical contact with objects and maintain balance. The operator does not need to wear any tracking device. The robot is purely driven by the same RGB signal processing module (SEG) and tracking policy as in the dance experiment, without task-specific training. Two tasks are shown in Fig.~\ref{fig:teleop_manip}.

In Fig.~\ref{fig:teleop_manip}-(a), a box stands upright on a table in front of the robot, and another table stands on the robot's left. The robot clamps the box between both hands, lifts and tilts it, turns toward the side table, and lays the box down on its side before releasing it. The clamp holds only while both hands keep pressing on opposite sides of the box, so the relative configuration of the two arms must be preserved as the torso turns. The box remains clamped throughout the transfer, indicating that the references produced by reconstruction, retargeting, and extrapolation preserve this bimanual coordination during the turn. When the box rests tilted against the edge of the side table, the operator adjusts the arms before pushing it flat. Such corrections rely on the operator's visual feedback and on the policy following small corrective arm motions while the hands are in contact with the object.

Fig.~\ref{fig:teleop_manip}-(b) shows a continuous session that a bag initially rests on a blacktop small table. The robot hooks one hand through the bag handles, lifts the bag while stepping, carries it beside the stand, and rests it on the near small table, and finally withdraws its hand from the handles. During this withdrawal, the tracked image position of the bag remains unchanged within one pixel, indicating that the hand is extracted from the handle loop without dragging the bag. Because the policy observes neither the object nor the contact forces, this load is an unmodeled disturbance that must be compensated through proprioceptive feedback alone. The robot remained balanced, and the bag stayed hooked on its hand.

Compared with the Bencaogangmu dance, these tasks involve slow, aperiodic, upper-body-dominated motions, while the lower body must stabilize the robot against object loads. Their completion suggests that the $\theta$-conditioned tracker and the reference generation are not limited to rhythmic motions. Neither the visual reconstruction nor the policy observes the object, and monocular reconstruction provides limited depth cues for such fine alignment. Incorporating object perception or haptic feedback is therefore a natural extension of this interface.

\section{Conclusion and Discussion}
We presented MoViC, a framework for humanoid whole-body control from live monocular RGB video. The frequency-domain motion representations supply temporal context with fewer policy input features than normal design, while a single next-step target retains direct tracking guidance. This separation gives the transformer complementary information about the immediate tracking objective and the surrounding motion, providing a structured way to use temporal information in the reference library. Our extrapolation method enables 50-Hz reference generation between slower visual updates, and measurement anchoring compensates for shared reconstruction offsets. The asynchronous pipeline allows visual reconstruction and policy execution to operate at different rates. This avoids forcing the controller to wait for every visual update, while retaining an input interface based on a single RGB camera. Hardware deployment demonstrates that the framework can track highly dynamic dance motions from live video. The same tracking policy also accomplishes device-free teleoperation tasks, including bimanual and single-arm object manipulation, without task-specific training

These advantages come with trade-offs. Reference quality still depends on visual reconstruction and retargeting, and latent phase propagation cannot guarantee accurate predictions during prolonged occlusion or abrupt motion changes. Future work will focus on improving the accuracy and processing speed of video-based human motion reconstruction. We will also extend the current fixed-camera setup to support egocentric video from a camera mounted on the robot's head, accounting for camera motion and viewpoint changes during robot movement. More motion data should also be incorporated to discover the scaling law of the frequency-domain representations. The frequency-domain representation summarizes local temporal structure in a shared feature space, providing a compact conditioning interface across diverse motion clips. This structure may facilitate the reuse of temporal features across clips. We hypothesize that such conditioning can help the tracking policy benefit more effectively from increases in the size and diversity of the motion dataset.

\bibliographystyle{IEEEtran}
\bibliography{root}

@inproceedings{he2025hover,
  title={Hover: Versatile neural whole-body controller for humanoid robots},
  author={He, Tairan and Xiao, Wenli and Lin, Toru and Luo, Zhengyi and Xu, Zhenjia and Jiang, Zhenyu and Kautz, Jan and Liu, Changliu and Shi, Guanya and Wang, Xiaolong and others},
  booktitle={2025 IEEE International Conference on Robotics and Automation (ICRA)},
  pages={9989--9996},
  year={2025},
  organization={IEEE}
}

@inproceedings{he2024learning,
  title={Learning human-to-humanoid real-time whole-body teleoperation},
  author={He, Tairan and Luo, Zhengyi and Xiao, Wenli and Zhang, Chong and Kitani, Kris and Liu, Changliu and Shi, Guanya},
  journal={arXiv preprint arXiv:2403.04436},
  year={2024}
}

@article{he2024omnih2o,
  title={Omnih2o: Universal and dexterous human-to-humanoid whole-body teleoperation and learning},
  author={He, Tairan and Luo, Zhengyi and He, Xialin and Xiao, Wenli and Zhang, Chong and Zhang, Weinan and Kitani, Kris and Liu, Changliu and Shi, Guanya},
  journal={arXiv preprint arXiv:2406.08858},
  year={2024}
}

@article{ze2025twist,
title={TWIST: Teleoperated Whole-Body Imitation System},
author= {Yanjie Ze and Zixuan Chen and João Pedro Araújo and Zi-ang Cao and Xue Bin Peng and Jiajun Wu and C. Karen Liu},
year= {2025},
journal= {arXiv preprint arXiv:2505.02833}
}

@article{mao2024learning,
  title={Learning from Massive Human Videos for Universal Humanoid Pose Control},
  author={Mao, Jiageng and Zhao, Siheng and Song, Siqi and Shi, Tianheng and Ye, Junjie and Zhang, Mingtong and Geng, Haoran and Malik, Jitendra and Guizilini, Vitor and Wang, Yue},
  journal={arXiv preprint arXiv:2412.14172},
  year={2024}
}

@misc{weng2025hdmilearninginteractivehumanoid,
      title={HDMI: Learning Interactive Humanoid Whole-Body Control from Human Videos}, 
      author={Haoyang Weng and Yitang Li and Nikhil Sobanbabu and Zihan Wang and Zhengyi Luo and Tairan He and Deva Ramanan and Guanya Shi},
      year={2025},
      eprint={2509.16757},
      archivePrefix={arXiv},
      primaryClass={cs.RO},
      url={https://arxiv.org/abs/2509.16757}, 
}

@inproceedings{pavlakos2024reconstructing,
  title={Reconstructing Hands in 3{D} with Transformers},
  author={Pavlakos, Georgios and Shan, Dandan and Radosavovic, Ilija and Kanazawa, Angjoo and Fouhey, David and Malik, Jitendra},
  booktitle={CVPR},
  year={2024}
}

@article{ze2025twist2,
  title={Twist2: Scalable, portable, and holistic humanoid data collection system},
  author={Ze, Yanjie and Zhao, Siheng and Wang, Weizhuo and Kanazawa, Angjoo and Duan, Rocky and Abbeel, Pieter and Shi, Guanya and Wu, Jiajun and Liu, C Karen},
  journal={arXiv preprint arXiv:2511.02832},
  year={2025}
}

@article{luo2026sonic,
  title={Sonic: Supersizing motion tracking for natural humanoid whole-body control},
  author={Luo, Zhengyi and Yuan, Ye and Wang, Tingwu and Li, Chenran and Casta{\~n}eda, Fernando and Chen, Sirui and Cao, Zi-Ang and Li, Jiefeng and Minor, David and Ben, Qingwei and others},
  journal={Science Robotics},
  volume={11},
  number={117},
  pages={eaed4592},
  year={2026},
  publisher={American Association for the Advancement of Science}
}

@inproceedings{li2024fld,
  title={Fld: Fourier latent dynamics for structured motion representation and learning},
  author={Li, Chenhao and Stanger-Jones, Elijah and Heim, Steve and others},
  booktitle={International Conference on Learning Representations},
  volume={2024},
  pages={53148--53174},
  year={2024}
}

@inproceedings{watanabe2025dfm,
  title={Dfm: Deep fourier mimic for expressive dance motion learning},
  author={Watanabe, Ryo and Li, Chenhao and Hutter, Marco},
  booktitle={2025 IEEE International Conference on Robotics and Automation (ICRA)},
  pages={9644--9650},
  year={2025},
  organization={IEEE}
}

@inproceedings{yang2026sam,
  title={Sam 3d body: Robust full-body human mesh recovery},
  author={Yang, Xitong and Kukreja, Devansh and Pinkus, Don and Fan, Taosha and Park, Jinhyung and Shin, Soyong and Cao, Jinkun and Liu, Jia-Wei and Ugrinovic, Nicol{\'a}s and Sagar, Anushka and others},
  booktitle={Proceedings of the IEEE/CVF Conference on Computer Vision and Pattern Recognition},
  pages={7209--7219},
  year={2026}
}

@article{shen2026gvhmr,
  title={GVHMR: Gravity-View Coordinates for Global Human Motion Recovery From Monocular Videos},
  author={Shen, Zehong and Wang, Jiawei and Pi, Huaijin and Xia, Yan and Cen, Zhi and Hu, Zechen and Bao, Hujun and Hu, Ruizhen and Zhou, Xiaowei and Peng, Sida},
  journal={IEEE Transactions on Pattern Analysis and Machine Intelligence},
  year={2026},
  publisher={IEEE}
}

@inproceedings{shin2024wham,
  title={Wham: Reconstructing world-grounded humans with accurate 3d motion},
  author={Shin, Soyong and Kim, Juyong and Halilaj, Eni and Black, Michael J},
  booktitle={2024 IEEE/CVF Conference on Computer Vision and Pattern Recognition (CVPR)},
  pages={2070--2080},
  year={2024},
  organization={IEEE}
}

@inproceedings{mahmood2019amass,
  title={AMASS: Archive of motion capture as surface shapes},
  author={Mahmood, Naureen and Ghorbani, Nima and Troje, Nikolaus F and Pons-Moll, Gerard and Black, Michael J},
  booktitle={Proceedings of the IEEE/CVF international conference on computer vision},
  pages={5442--5451},
  year={2019}
}

@article{he2025asap,
  title={Asap: Aligning simulation and real-world physics for learning agile humanoid whole-body skills},
  author={He, Tairan and Gao, Jiawei and Xiao, Wenli and Zhang, Yuanhang and Wang, Zi and Wang, Jiashun and Luo, Zhengyi and He, Guanqi and Sobanbab, Nikhil and Pan, Chaoyi and others},
  journal={arXiv preprint arXiv:2502.01143},
  year={2025}
}

@article{liao2026beyondmimic,
  title={Beyondmimic: From motion tracking to versatile humanoid control via guided diffusion},
  author={Liao, Qiayuan and Truong, Takara E and Huang, Xiaoyu and Gao, Yuman and Tevet, Guy and Sreenath, Koushil and Liu, C Karen},
  journal={Science Robotics},
  volume={11},
  number={117},
  pages={eadx8924},
  year={2026},
  publisher={American Association for the Advancement of Science}
}

@inproceedings{kocabas2020vibe,
  author    = {Kocabas, M. and Athanasiou, N. and Black, M. J.},
  title     = {{VIBE}: Video inference for human body pose and shape estimation},
  booktitle = {Proc. IEEE/CVF Conf. Computer Vision and Pattern Recognition (CVPR)},
  year      = {2020},
  pages     = {5253--5263},
  url       = {https://openaccess.thecvf.com/content_CVPR_2020/html/Kocabas_VIBE_Video_Inference_for_Human_Body_Pose_and_Shape_Estimation_CVPR_2020_paper.html}
}

@inproceedings{goel2023humans4d,
  author    = {Goel, S. and Pavlakos, G. and Rajasegaran, J. and Kanazawa, A. and Malik, J.},
  title     = {Humans in {4D}: Reconstructing and tracking humans with transformers},
  booktitle = {Proc. IEEE/CVF Int. Conf. Computer Vision (ICCV)},
  year      = {2023},
  pages     = {14783--14794},
  url       = {https://arxiv.org/abs/2305.20091}
}

@article{vandermaaten2008tsne,
  author  = {van der Maaten, Laurens and Hinton, Geoffrey},
  title   = {Visualizing data using {t-SNE}},
  journal = {Journal of Machine Learning Research},
  volume  = {9},
  number  = {86},
  pages   = {2579--2605},
  year    = {2008},
  url     = {https://jmlr.org/papers/v9/vandermaaten08a.html}
}

@article{chen2025gmt,
  title={Gmt: General motion tracking for humanoid whole-body control},
  author={Chen, Zixuan and Ji, Mazeyu and Cheng, Xuxin and Peng, Xuanbin and Peng, Xue Bin and Wang, Xiaolong},
  journal={arXiv preprint arXiv:2506.14770},
  year={2025}
}

@article{yin2026unitracker,
  title={Unitracker: Learning universal whole-body motion tracker for humanoid robots},
  author={Yin, Kangning and Zeng, Weishuai and Fan, Ke and Dai, Minyue and Wang, Zirui and Zhang, Qiang and Tian, Zheng and Wang, Jingbo and Pang, Jiangmiao and Zhang, Weinan},
  journal={IEEE Robotics and Automation Letters},
  year={2026},
  publisher={IEEE}
}

@inproceedings{tang2024humanmimic,
  title={Humanmimic: Learning natural locomotion and transitions for humanoid robot via wasserstein adversarial imitation},
  author={Tang, Annan and Hiraoka, Takuma and Hiraoka, Naoki and Shi, Fan and Kawaharazuka, Kento and Kojima, Kunio and Okada, Kei and Inaba, Masayuki},
  booktitle={2024 IEEE International Conference on Robotics and Automation (ICRA)},
  pages={13107--13114},
  year={2024},
  organization={IEEE}
}

@article{peng2018deepmimic,
  title={Deepmimic: Example-guided deep reinforcement learning of physics-based character skills},
  author={Peng, Xue Bin and Abbeel, Pieter and Levine, Sergey and Van de Panne, Michiel},
  journal={ACM Transactions On Graphics (TOG)},
  volume={37},
  number={4},
  pages={1--14},
  year={2018},
  publisher={ACM New York, NY, USA}
}

@article{peng2021amp,
  title={Amp: Adversarial motion priors for stylized physics-based character control},
  author={Peng, Xue Bin and Ma, Ze and Abbeel, Pieter and Levine, Sergey and Kanazawa, Angjoo},
  journal={ACM Transactions on Graphics (ToG)},
  volume={40},
  number={4},
  pages={1--20},
  year={2021},
  publisher={ACM New York, NY, USA}
}

@article{qi2026humanoid,
  title={Humanoid-GPT: Scaling Data and Structure for Zero-Shot Motion Tracking},
  author={Qi, Zekun and Chen, Xuchuan and Liu, Dairu and Lin, Chenghuai and Lian, Yunrui and Liang, Sikai and Zhang, Zhikai and Guan, Yu and Wang, Jilong and Zhang, Wenyao and others},
  journal={arXiv preprint arXiv:2606.03985},
  year={2026}
}

@article{wang2026experts,
  title={From experts to a generalist: Toward general whole-body control for humanoid robots},
  author={Wang, Yuxuan and Yang, Ming and Ding, Gang and Zhang, Yu and Zeng, Weishuai and Xu, Xinrun and Jiang, Haobin and Lu, Zongqing},
  journal={Advances in Neural Information Processing Systems},
  volume={38},
  pages={147748--147772},
  year={2026}
}

@article{tessler2024maskedmimic,
  title={Maskedmimic: Unified physics-based character control through masked motion inpainting},
  author={Tessler, Chen and Guo, Yunrong and Nabati, Ofir and Chechik, Gal and Peng, Xue Bin},
  journal={ACM Transactions On Graphics (TOG)},
  volume={43},
  number={6},
  pages={1--21},
  year={2024},
  publisher={ACM New York, NY, USA}
}

@inproceedings{luo2023perpetual,
  title={Perpetual humanoid control for real-time simulated avatars},
  author={Luo, Zhengyi and Cao, Jinkun and Kitani, Kris and Xu, Weipeng and others},
  booktitle={Proceedings of the IEEE/CVF International Conference on Computer Vision},
  pages={10895--10904},
  year={2023}
}

@inproceedings{SMPL-X:2019,
  title = {Expressive Body Capture: {3D} Hands, Face, and Body from a Single Image},
  author = {Pavlakos, Georgios and Choutas, Vasileios and Ghorbani, Nima and Bolkart, Timo and Osman, Ahmed A. A. and Tzionas, Dimitrios and Black, Michael J.},
  booktitle = {Proceedings IEEE Conf. on Computer Vision and Pattern Recognition (CVPR)},
  pages     = {10975--10985},
  year = {2019}
}

@article{fu2024humanplus,
  title={{HumanPlus}: Humanoid Shadowing and Imitation from Humans},
  author={Fu, Zipeng and Zhao, Qingqing and Wu, Qi and Wetzstein, Gordon and Finn, Chelsea},
  journal={arXiv preprint arXiv:2406.10454},
  year={2024},
  url={https://arxiv.org/abs/2406.10454}
}

@inproceedings{tessler2023calm,
  title={Calm: Conditional adversarial latent models for directable virtual characters},
  author={Tessler, Chen and Kasten, Yoni and Guo, Yunrong and Mannor, Shie and Chechik, Gal and Peng, Xue Bin},
  booktitle={ACM SIGGRAPH 2023 conference proceedings},
  pages={1--9},
  year={2023}
}

@article{joao2025gmr,
title={Retargeting Matters: General Motion Retargeting for Humanoid Motion Tracking},
author= {Joao Pedro Araujo and Yanjie Ze and Pei Xu and Jiajun Wu and C. Karen Liu},
year= {2025},
journal= {arXiv preprint arXiv:2510.02252}
}

@article{zhu2023neural,
  title={Neural categorical priors for physics-based character control},
  author={Zhu, Qingxu and Zhang, He and Lan, Mengting and Han, Lei},
  journal={ACM Transactions on Graphics (TOG)},
  volume={42},
  number={6},
  pages={1--16},
  year={2023},
  publisher={ACM New York, NY, USA}
}

@article{chen2026holomotion,
  title={Holomotion-1 technical report},
  author={Chen, Maiyue and Wang, Kaihui and Zhang, Bo and Ma, Xihan and Yang, Zhiyuan and Ren, Yi and Huang, Qijun and Zhu, Zihao and Wang, Yucheng and Su, Zhizhong},
  journal={arXiv preprint arXiv:2605.15336},
  year={2026}
}

@article{tao2026heracles,
  title={Heracles: Bridging Precise Tracking and Generative Synthesis for General Humanoid Control},
  author={Tao, Zelin and Su, Zeran and Liu, Peiran and Sun, Jingkai and Que, Wenqiang and Ma, Jiahao and Yu, Jialin and Cao, Jiahang and Sun, Pihai and Liang, Hao and others},
  journal={arXiv preprint arXiv:2603.27756},
  year={2026}
}

@article{yin2025visualmimic,
  title={Visualmimic: Visual humanoid loco-manipulation via motion tracking and generation},
  author={Yin, Shaofeng and Ze, Yanjie and Yu, Hong-Xing and Liu, C Karen and Wu, Jiajun},
  journal={arXiv preprint arXiv:2509.20322},
  year={2025}
}

@article{ni2025generated,
  title={From generated human videos to physically plausible robot trajectories},
  author={Ni, James and Wang, Zekai and Lin, Wei and Bar, Amir and LeCun, Yann and Darrell, Trevor and Malik, Jitendra and Herzig, Roei},
  journal={arXiv preprint arXiv:2512.05094},
  year={2025}
}

@article{starke2022deepphase,
  title={Deepphase: Periodic autoencoders for learning motion phase manifolds},
  author={Starke, Sebastian and Mason, Ian and Komura, Taku},
  journal={ACM Transactions on Graphics (ToG)},
  volume={41},
  number={4},
  pages={1--13},
  year={2022},
  publisher={ACM New York, NY, USA}
}

@article{harvey2020robust,
author    = {Félix G. Harvey and Mike Yurick and Derek Nowrouzezahrai and Christopher Pal},
title     = {Robust Motion In-Betweening},
booktitle = {ACM Transactions on Graphics (Proceedings of ACM SIGGRAPH)},
publisher = {ACM},
volume    = {39},
number    = {4},
year      = {2020}
}

@article{lin2023motionx,
  title={Motion-X: A Large-scale 3D Expressive Whole-body Human Motion Dataset},
  author={Lin, Jing and Zeng, Ailing and Lu, Shunlin and Cai, Yuanhao and Zhang, Ruimao and Wang, Haoqian and Zhang, Lei},
  journal={Advances in Neural Information Processing Systems},
  year={2023}
}

@article{li2023object,
  title={Object Motion Guided Human Motion Synthesis},
  author={Li, Jiaman and Wu, Jiajun and Liu, C Karen},
  journal={ACM Trans. Graph.},
  volume={42},
  number={6},
  year={2023}
}

@article{omnitrack2026,
  title   = {OmniTrack: General Motion Tracking via Physics-Consistent Reference},
  author  = {Yuhan, Li and et al.},
  journal = {arXiv preprint arXiv:2602.23832},
  year    = {2026}
}

@article{clevert2015fast,
  title={Fast and accurate deep network learning by exponential linear units (elus)},
  author={Clevert, Djork-Arn{\'e} and Unterthiner, Thomas and Hochreiter, Sepp},
  journal={arXiv preprint arXiv:1511.07289},
  year={2015}
}

@article{simeoni2025dinov3,
  title={Dinov3},
  author={Sim{\'e}oni, Oriane and Vo, Huy V and Seitzer, Maximilian and Baldassarre, Federico and Oquab, Maxime and Jose, Cijo and Khalidov, Vasil and Szafraniec, Marc and Yi, Seungeun and Ramamonjisoa, Micha{\"e}l and others},
  journal={arXiv preprint arXiv:2508.10104},
  year={2025}
}

@misc{wang2025moge2accuratemonoculargeometry,
      title={MoGe-2: Accurate Monocular Geometry with Metric Scale and Sharp Details}, 
      author={Ruicheng Wang and Sicheng Xu and Yue Dong and Yu Deng and Jianfeng Xiang and Zelong Lv and Guangzhong Sun and Xin Tong and Jiaolong Yang},
      year={2025},
      eprint={2507.02546},
      archivePrefix={arXiv},
      primaryClass={cs.CV},
      url={https://arxiv.org/abs/2507.02546}, 
}

@article{ferguson2025mhr,
  title={Mhr: Momentum human rig},
  author={Ferguson, Aaron and Osman, Ahmed AA and Bescos, Berta and Stoll, Carsten and Twigg, Chris and Lassner, Christoph and Otte, David and Vignola, Eric and Prada, Fabian and Bogo, Federica and others},
  journal={arXiv preprint arXiv:2511.15586},
  year={2025}
}

@article{ji2024exbody2,
  title={ExBody2: Advanced Expressive Humanoid Whole-Body Control}, 
  author={Ji, Mazeyu and Peng, Xuanbin and Liu, Fangchen and Li, Jialong and Yang, Ge and Cheng, Xuxin and Wang, Xiaolong},
  journal={arXiv preprint arXiv:2412.13196},
  year={2024},
  }

@misc{li2025clone,
          title={CLONE: Closed-Loop Whole-Body Humanoid Teleoperation for Long-Horizon Tasks}, 
          author={Yixuan Li and Yutang Lin and Jieming Cui and Tengyu Liu and Wei Liang and Yixin Zhu and Siyuan Huang},
          journal={arXiv preprint arXiv:2506.08931}, 
          year={2025}
        }

@inproceedings{
luo2024universal,
title={Universal Humanoid Motion Representations for Physics-Based Control},
author={Zhengyi Luo and Jinkun Cao and Josh Merel and Alexander Winkler and Jing Huang and Kris M. Kitani and Weipeng Xu},
booktitle={The Twelfth International Conference on Learning Representations},
year={2024},
url={https://openreview.net/forum?id=OrOd8PxOO2}
}

@article{schulman2017proximal,
  title={Proximal policy optimization algorithms},
  author={Schulman, John and Wolski, Filip and Dhariwal, Prafulla and Radford, Alec and Klimov, Oleg},
  journal={arXiv preprint arXiv:1707.06347},
  year={2017}
}

@inproceedings{li2021hybrik,
    title={Hybrik: A hybrid analytical-neural inverse kinematics solution for 3d human pose and shape estimation},
    author={Li, Jiefeng and Xu, Chao and Chen, Zhicun and Bian, Siyuan and Yang, Lixin and Lu, Cewu},
    booktitle={Proceedings of the IEEE/CVF Conference on Computer Vision and Pattern Recognition},
    pages={3383--3393},
    year={2021}
}

@inproceedings{ye2023slahmr,
    title={Decoupling Human and Camera Motion from Videos in the Wild},
    author={Ye, Vickie and Pavlakos, Georgios and Malik, Jitendra and Kanazawa, Angjoo},
    booktitle={IEEE Conference on Computer Vision and Pattern Recognition (CVPR)},
    month={June},
    year={2023}
}

@inproceedings{mono-3dhp2017,
 author = {Mehta, Dushyant and Rhodin, Helge and Casas, Dan and Fua, Pascal and Sotnychenko, Oleksandr and Xu, Weipeng and Theobalt, Christian},
 title = {Monocular 3D Human Pose Estimation In The Wild Using Improved CNN Supervision},
 booktitle = {3D Vision (3DV), 2017 Fifth International Conference on},
 url = {http://gvv.mpi-inf.mpg.de/3dhp_dataset},
 year = {2017},
 organization={IEEE},
 doi={10.1109/3dv.2017.00064},
}

@inproceedings{kaufmann2023emdb,
  author = {Kaufmann, Manuel and Song, Jie and Guo, Chen and Shen, Kaiyue and Jiang, Tianjian and Tang, Chengcheng and Z{\'a}rate, Juan Jos{\'e} and Hilliges, Otmar},
  title = {{EMDB}: The {E}lectromagnetic {D}atabase of {G}lobal 3{D} {H}uman {P}ose and {S}hape in the {W}ild},
  booktitle = {International Conference on Computer Vision (ICCV)},
  year = {2023}
}

\end{document}